%% file: 0_pcc.tex
\documentclass[journal,compsoc]{IEEEtran}

\usepackage{cite}
\usepackage{cite}\usepackage[pagebackref=true,breaklinks=true,colorlinks,bookmarks=false]{hyperref} 
\usepackage{url}   
\usepackage{float}
\usepackage{graphicx}
\usepackage{subfig}
\usepackage{amsmath}
\usepackage{array}
\usepackage{multirow}
\usepackage{flushend}
\usepackage{amssymb}
\usepackage{amsfonts}
\usepackage{booktabs}
\usepackage{textcomp}
\usepackage{soul} 
\usepackage{ulem}

\begin{document}
%
\title{Sparse Tensor-based Multiscale Representation for Point Cloud Geometry Compression}
%
%
\author{Jianqiang Wang, 
        Dandan Ding,  
        Zhu Li, 
        Xiaoxing Feng,
        Chuntong Cao,
        and Zhan Ma\\
\thanks{J. Wang and Z. Ma are both with the School of Electronic Science and Engineering, Nanjing University, Nanjing, Jiangsu 210093, China. Emails: wangjq@smail.nju.edu.cn, mazhan@nju.edu.cn.}
\thanks{D. Ding is with HangZhou Normal University, Hangzhou, Zhejiang, China. Email: dandanding@hznu.edu.cn}
\thanks{Z. Li is with the University of Missouri, Kansas City, MO 64110, USA. Email: zhu.li@ieee.org.}
\thanks{X. Feng and C. Cao are with the Jiangsu Longyuan Zhenhua Marine Engineering Co., LTD., Nantong, Jiangsu, China. Emails: fengxiaoxing@zpmc.com, 25519785@qq.com}
}

%



\IEEEtitleabstractindextext{
\begin{abstract}
This study develops a unified Point Cloud Geometry (PCG) compression method through the processing of multiscale sparse tensor-based voxelized PCG. 
We call this compression method SparsePCGC.
The proposed SparsePCGC is a low complexity solution because it only performs the convolutions on sparsely-distributed Most-Probable Positively-Occupied Voxels (MP-POV).  
The multiscale representation also allows us to compress scale-wise MP-POVs by exploiting cross-scale and same-scale correlations extensively and flexibly.
The overall compression efficiency highly depends on the accuracy of estimated occupancy probability for each MP-POV. Thus, we first design the Sparse Convolution-based Neural Network (SparseCNN) which stacks sparse convolutions and voxel sampling to best characterize and embed spatial correlations. 
We then develop the SparseCNN-based Occupancy Probability Approximation (SOPA) model to estimate the occupancy probability either in a single-stage manner only using the cross-scale correlation, or in a multi-stage manner by exploiting stage-wise correlation among same-scale neighbors.
Besides, we also suggest the SparseCNN based Local Neighborhood Embedding (SLNE) to aggregate local variations as spatial priors in feature attribute to improve the SOPA.
Our unified approach not only shows state-of-the-art performance in both lossless and lossy compression modes across a variety of datasets including the dense object PCGs (8iVFB, Owlii, MUVB) and sparse LiDAR PCGs (KITTI, Ford) when compared with standardized MPEG G-PCC and other prevalent learning-based schemes, but also has low complexity which is attractive to practical applications. 
\end{abstract}

\begin{IEEEkeywords}
Point cloud geometry compression, Sparse Tensor, Sparse convolution,  Multiscale Representation, Occupancy Probability Approximation, Neighborhood Embedding 
\end{IEEEkeywords}
}

\maketitle
%

\input{1_intro}
\input{2_related}
\input{3_method_part1}

\input{3_method_part2}

\input{4_exp_part1}

\input{4_exp_part2}

\input{5_conclusion}

\bibliographystyle{IEEEtran}
\bibliography{pccbib}
\end{document}

%% file: 1_intro.tex
\section{Introduction}
\label{sec:intro}
Due to their outstanding flexibility for representing 3D objects realistically and naturally, {point clouds} have become a popular media format used in a large number of applications, such as the Augmented/Virtual Reality (AR/VR), autonomous driving, and cultural e-heritage. This then raises an urgent need for high-efficiency lossless and lossy compression of point clouds~\cite{schwarz2019emerging}. This work first exemplifies the development of {\it lossless compression} in detail, since existing industrial applications demand such functionality for archiving point cloud-based digital assets efficiently, such as the 3D building information models used in digital twin~\cite{digitalTwin}, and for enforcing safety-critical system using high-precision 3D point cloud maps in autonomous robots. Later then, we show that the same architecture can be easily extended for {\it lossy compression} as well.

A point cloud is a collection of non-uniformly and sparsely distributed points that can be characterized using their 3D coordinates (e.g., $(x,y,z)$) and  attributes (e.g., RGB colors, reflectances) if applicable. Unlike those well-structured pixel grids of a 2D image plane or a video frame, a point cloud relies on unconstrained displacement of points to represent arbitrary-shaped 3D objects flexibly. However, this creates issues for the efficient coding of geometric occupancy due to difficulties in characterizing and exploiting inter-correlation across irregularly scattered points in a free 3D space. The main focus of this paper is Point Cloud Geometry (PCG) compression. For the sake of simplicity, we start our discussion from the lossless mode of dense object Point Cloud Geometry Compression (PCGC), and then extend the studies to other scenarios.

\begin{figure}[t]
	\begin{center}
	\includegraphics[width=3in]{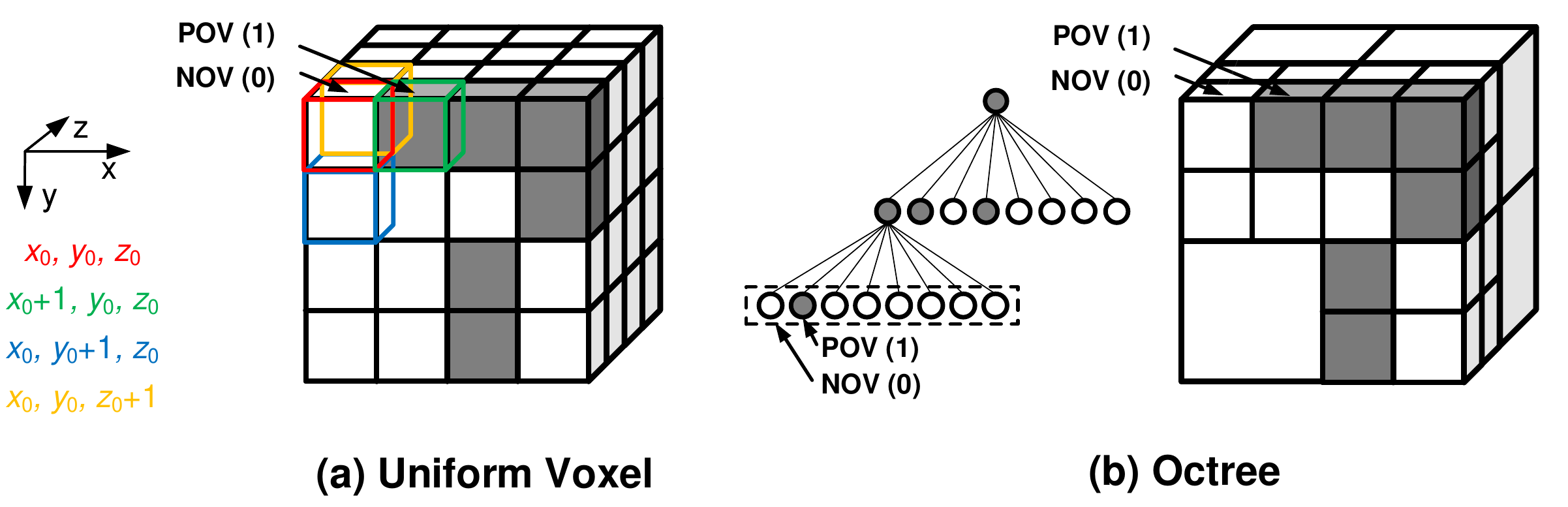}
	\includegraphics[width=3.3in]{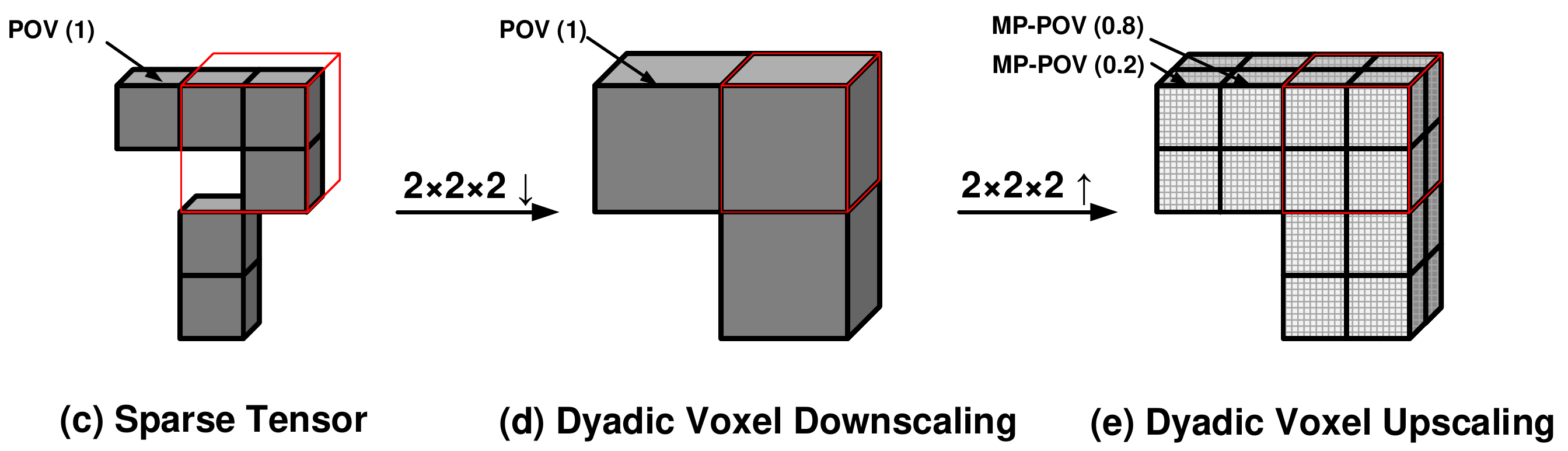}
	\end{center}
	\caption{{\bf Various Representation Models of Voxelized PCG}: (a) uniform voxel; (b)  octree model (and parent-tree layout); (c) sparse tensor; (d) dyadic voxel downscaling  $2\times2\times2\downarrow$ from (c); (e) dyadic voxel upscaling $2\times2\times2\uparrow$ from (d) to generate MP-POVs and associated occupancy probability, i.e., MP-POV($p_{\text{MP-POV}}$). Voxel grids with the solid line are all involved in computation. {\it Solid grey voxels are positively occupied, a.k.a., POVs; and the rest are NOVs, in (a) and (b). The red box in (c), (d), and (e) contains voxels used in resampling. Voxels painted with dense grid pattern are MP-POVs.} }
	\label{fig:data_strucutre}
\end{figure}

\subsection{Motivation}
Usually, the compression efficiency of a point in a point cloud is closely related to its probability approximation conditioned on available neighbors~\cite{CABAC,InfoTheory_bible}. The entropy $R_{\vec{v}_k}$ {after compression} can be represented as: 
\begin{align}
    R_{\vec{v}_k} = \mathbb{E}\left[-\log_2p_{\vec{v}}({\vec{v}_k}|\vec{v}_{k-1},\vec{v}_{k-2}\ldots \vec{v}_{k_0})\right]. \label{eq:cond_context_modeling}
\end{align} Here, $\vec{v}_k$ is the $k$-th to-be-compressed point, and $\vec{v}_{k-1}$, $\vec{v}_{k-2}$, $\ldots \vec{v}_{k_0}$ are its causal neighbors that are already processed and served as the prior knowledge for the process of current $k$-th element. $p_{\vec{v}}({\vec{v}_k}|\vec{v}_{k-1},\vec{v}_{k-2}\ldots \vec{v}_{k_0})$ is the conditional probability of $\vec{v}_k$ which is determined by a context model and used to approximate $R_{\vec{v}_k}$. The more precise the context modeling is, the closer the probability distribution to the real data is, and the fewer bits are consumed.

To build adaptive contexts using spatial neighbors, a number of 3D representation models have been developed, including the ``uniform voxel model'' in Fig.~\ref{fig:data_strucutre}a and ``octree model'' in Fig.~\ref{fig:data_strucutre}b~\cite{meagher1982geometric}, on top of which either rules-based~\cite{schwarz2019emerging,MPEG_PCC_PIEEE} or deep neural network (DNN)-based context models~\cite{quach2019learning,Wang2021Lossy,Guarda2021AdaptiveDL,Nguyen2021MultiscaleDC,Huang2020OctSqueezeOE,NEURIPS2020,Que2021VoxelContextNetAO} can be devised to explore correlations across spatial neighbors.

\begin{figure*}[t]
\centering
\includegraphics[width=7in]{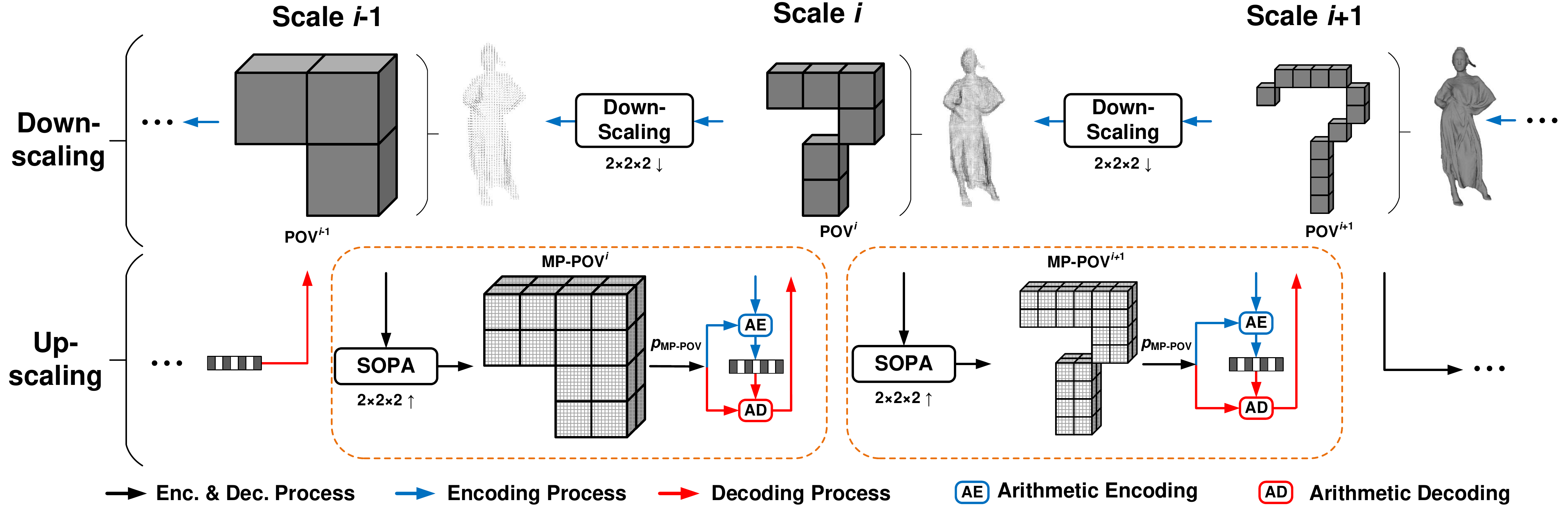}\label{fig:multiscale}
\caption{{\bf SparsePCGC.} In the encoding process (blue arrows), we progressively downscale the {PCG tensor} and encode {occupied voxels} at each scale into the binary stream according to their occupancy probability; As for the decoding flow (red arrows), we exactly reconstruct occupied voxels, e.g., POV or NOV, by decoding the binary bits with the occupancy probability. 
The key component, a.k.a., occupancy probability approximation, that plays a vital role for compression efficiency is developed by utilizing the cross-scale and same-scale priors and is plugged in both encoder and decoder for bit-wise match. Such SparseCNN-based Occupancy Probability Approximation is referred to as the ``SOPA''. The SOPA can be fulfilled through either single-stage or multi-stage manner, depending on the tradeoff between the performance and complexity  in applications.}
\label{fig:overview}
\end{figure*}

Although those DNN models improve the compression efficiency noticeably against rules-based approaches like recently-approved ISO/IEC MPEG (Moving Picture Experts Group) Geometry-based PCC (G-PCC) standard~\cite{schwarz2019emerging,MPEG_PCC_PIEEE}, their huge complexity hinders the prompt use of these models in practice. On the other hand,  alternative point-wise processing methods\footnote{Note that aforementioned uniform voxel representation and octree decomposition do require the voxelization to pre-process raw input, while currently these point-wise methods do not necessarily need this step. }~\cite{Huang20193DPC, Gao2021PCGCGraphSampling,Qi2017PointNetDH} like PointNet++ in~\cite{Qi2017PointNetDH} report low complexity, but their compression performance largely suffers~\cite{Gao2021PCGCGraphSampling}. 
More importantly, these learning-based approaches cannot easily generalize themselves to various compression scenarios, such as the support of both {dense and sparse point clouds} and the support of both lossy and lossless modes. 

After the conclusion of G-PCC~\cite{MPEG_PCC_PIEEE}, international standardization groups, such as the ISO/IEC MPEG, and ISO/IEC JPEG (Joint Photographic Experts Group) are actively pursuing the next-generation point cloud compression using learning-based approaches~\cite{JPEG2019CTC,MPEG_LearntPCC}, which urges a solution with {\it better coding efficiency, affordable complexity}, and {\it{robust} model generalization} at the same time. Towards this goal, a more efficient 3D representation method and a better spatial neighbor utilization mechanism for adaptive contexts  are highly desired.

\subsection{Our Approach}
Following the conventions used in standardized G-PCC~\cite{schwarz2019emerging} and other prevalent learning-based PCGC methods~\cite{Wang2021Lossy,Guarda2021AdaptiveDL,quach2019learning}, this study uses voxelized PCG for processing. A volumetric presentation of a PCG is depicted in Fig.~\ref{fig:data_strucutre}a using a densely-sampled uniform voxel grid  which is referred to as the ``uniform voxel'' representation for convenience. 
We then use {\it binary occupancy status} (or occupancy status) to describe whether the geometric position of the current voxel is positively occupied or not. For those Positively-Occupied Voxels (POVs) painted in solid grey in Fig.~\ref{fig:data_strucutre}, they are marked using indicators ``1'', e.g., POV(1); Conversely, Non-Occupied (empty) Voxels (NOVs) in white box  are indicated using  ``0'', e.g., NOV(0).  Apparently, those POVs are converted from raw ``points'' of an original PCG by voxelization.

\subsubsection{Multiscale Sparse Tensor Representation of PCG}

Facing those non-uniformly distributed POVs of an input PCG, this paper suggests the use of {\it Sparse Tensor} to best exploit the sparsity of POVs~\cite{BMVC2015_150}. Through the use of {sparse tensor}, we only cache the geometric coordinates and attributes (if applicable) for Most-Probable POVs, i.e., MP-POVs. 
These MP-POVs are generated from POVs of preceding lower scale via voxel upscaling, as  illustrated from Fig.~\ref{fig:data_strucutre}d to~\ref{fig:data_strucutre}e, to form scale-wise sparse tensor under a {\it multiscale representation} framework.

With this aim, we progressively downscale the original PCG into multi-resolution PCGs and upscale them correspondingly for hierarchical reconstruction, as  in Fig.~\ref{fig:overview}. We simply use {\it dyadic voxel sampling} to upscale or downscale related sparse tensor, which basically performs the scaling with a stride of 2 at each axis (i.e., $x$-, $y$- and $z$-axis in a Cartesian coordinate system). We regard the $2\times2\times2\uparrow$ (or $2^3\uparrow$) as dyadic voxel upscaling, and  $2\times2\times2\downarrow$ (or $2^3\downarrow$) as dyadic voxel downscaling.
\begin{itemize}
    \item As for the downscaling in encoding, we merge eight connected sub-voxels  in a $2\times2\times2$ voxelized box to a single voxel (from Fig.~\ref{fig:data_strucutre}c to~\ref{fig:data_strucutre}d). If any one out of these eight sub-voxels is a POV, the merged voxel is a POV. In the encoder, we have full knowledge of each POV at each scale.
    \item As for the upscaling in decoding, we sub-divide a specific voxel to eight sub-voxels  (from Fig.~\ref{fig:data_strucutre}d to~\ref{fig:data_strucutre}e). If this voxel is a POV\footnote{We know the ground truth after decoding in lossless compression mode because the sparse tensor in encoder and the decoded tensor in decoder have to be the same.}, its 8 children  are  marked as MP-POVs since we cannot decide which sub-voxel is positively occupied just right after the upscaling. The collection of  MP-POVs is a superset of  POVs.
\end{itemize}

\subsubsection{Cross-Scale Context Modeling}
The proposed SparsePCGC encodes the occupancy of MP-POVs at each scale into a compressed bitstream and correspondingly it decodes the binary bits to reconstruct same-scale POVs. For either lossy or lossless compression, the occupancy probability approximation of each MP-POV is critical to the performance of SparsePCGC.

Upon the multiscale representation model, we perform the cross-scale context modeling by utilizing decoded POVs (e.g., occupancy status and feature attribute if applicable) from the preceding lower scale to approximate the occupancy probability of each MP-POV at the current scale. 
Note that the cross-scale context modeling is limited between two consecutive scales only. This allows us to  train models just using samples from two consecutive scales and then apply them for inference across all scales. This cross-scale model sharing strategy reduces the model size significantly.

The proposed cross-scale context modeling is fulfilled by SparseCNN-based
Occupancy Probability Approximation (SOPA) that can be implemented in various ways. For example, we can approximate the occupancy probabilities of eight upscaled MP-POVs in one shot through the use of a single-stage SOPA.  We can also exploit neighborhood correlation  by applying multi-stage SOPA to progressively approximate the occupancy probabilities using previously-decoded POVs at the same scale from one group to another.

The aforementioned studies only explore correlations in the course of scale upsampling by applying the dyadic voxel downscaling in encoding without any spatial information embedding. Here we propose the SLNE (SparseCNN-based Local Neighborhood Embedding) to  characterize and embed local variations of each POV as its feature attribute in downscaling. Thereafter, when performing the scale upsampling,  both  occupancy status and latent features of each decoded POV are fed into SOPA model for better performance.

\subsection{Contribution}

\begin{table}[t]
    \centering
    \caption{Notations and Abbreviations}
    \label{tab:abbr}
    \begin{footnotesize}
    \begin{tabular}{c|c}
    \hline
    Abbreviation & Description\\
    \hline
    PCC & Point Cloud Compression\\
        PCG & Point Cloud Geometry\\
        PCGC & Point Cloud Geometry Compression\\
        POV & Positively-Occupied Voxel\\
        MP-POV & Most Probable POV\\
        NOV & Non-Occupied Voxel\\
        \hline
        \multirow{2}{*}{SparseCNN} & Sparse Convolution-based \\
        &  Neural Networks\\
        \hline
        \multirow{2}{*}{SOPA} & SparseCNN-based \\
        & Occupancy Probability Approximation\\
        \hline
        SOPA  & SparseCNN-based\\
        (Position) & Occupancy Position Adjustment\\
        \hline
        \multirow{2}{*}{SLNE} & SparseCNN-based \\
       & Local Neighborhood Embedding\\
     
        \hline

    G-PCC & Geometry-based PCC~\cite{tmc13code}\\
    \hline
    BD-Rate & Bj{\o}ntegaard Delta Rate~\cite{BDrate}\\
    \hline
    \end{tabular}
\end{footnotesize}
\end{table}

Our contributions are summarized as follows:
\begin{itemize} 
    \item Together with the preliminary exploration in~\cite{Wang2020MultiscalePC}, we are probably the {\it first one} to suggest the convolutional representation of multiscale sparse tensor for point cloud geometry compression, with which we effectively exploit correlations through cross-scale context modeling for better efficiency.
    \item The proposed SparsePCGC generalizes very well with state-of-the-art efficiency in both lossless and lossy scenarios across a variety of datasets, such as the densely-sampled 8i Voxelized Full Bodies (8iVFB)~\cite{8i20178i}, Owlii sequences~\cite{xu2017owlii} and Microsoft Voxelized Upper Bodies (MVUB)~\cite{microsoft2019microsoft}, and sparsely-sampled LiDAR sequences KITTI and Ford, in comparison to the standardized G-PCC and other learning-based approaches~\cite{Wang2021Lossy, Wang2020MultiscalePC,Quach2020ImprovedDP, Guarda2021AdaptiveDL,  Nguyen2021MultiscaleDC,Kaya2021NeuralNM,Huang2020OctSqueezeOE,Que2021VoxelContextNetAO,Fu2022OctAttentionOL}.
    \item Our method also demonstrates low space and time complexity. For example, its runtime for encoding and decoding is quantitatively comparable to the G-PCC, requiring just 1$\sim$2 seconds on average to encode/decode a large-scale PCG frame (see Table~\ref{table:models}), which is about orders of magnitude reduction compared to recently-published VoxelDNN~\cite{Nguyen2021LosslessCO},  NNOC~\cite{Kaya2021NeuralNM}, and OctAttention~\cite{Fu2022OctAttentionOL} (in decoding). Moreover, sharing the same  model across scales reduces the space complexity significantly, which is attractive to industrial practitioners.
\end{itemize} 

Table~\ref{tab:abbr} lists notations and abbreviations frequently used in this work for convenience. 

%% file: 2_related.tex
\section{Related Work} 
\label{sec:related_work}

This section reviews relevant techniques for the compression of PCG.

\subsection{Rules-based Approaches}
The octree model is the most popular format to represent voxelized point cloud~\cite{Jackins1980Oct}. A volumetric point cloud is recursively divided using octree decomposition until it reaches the leaf nodes/voxels. The occupancy of a voxel or sub-voxel can be then statistically compressed through predefined context prediction following the parent-child tree structure (see Fig.~\ref{fig:data_strucutre}b)~\cite{Schnabel2006Octree, Huang2008A}. This octree-based coding mechanism using handcrafted rules was adopted in MPEG G-PCC~\cite{schwarz2019emerging}, known as the {\it octree geometry codec}. 

To reconstruct object surfaces with finer structural details, a number of triangle meshes can be constructed locally on top of the octree model~\cite{graziosi2020overview}. MPEG G-PCC then included this triangulation-based mesh model, a.k.a., triangle soups representation of local geometry, into the test model, referred to as the {\it trisoup geometry codec}. 

Alternatively, traditional 2D image and video coding techniques~\cite{VVC_overview} have demonstrated superior compression efficiency and have been used in many applications. This motivates us to project a given 3D object to multiple 2D planes from a variety of viewpoints and then leverage popular image and video codecs for compact representation of projected 2D image sequences. Examples include the  MPEG  Video-based PCC~\cite{graziosi2020overview}, View-PCC~\cite{zhu2020view}, etc. Such 3D-to-2D projection based solutions explore a different route which is not the focus of this work.

\subsection{Learning-based Methods}

Numerous works have demonstrated that DNNs can best exploit inter-voxel correlations upon a specific 3D representation format for PCGC in either lossy or lossless mode. 

{\bf Uniform Voxel Representation.} Earlier attempts, including Wang~\textit{et al.}~\cite{Wang2021Lossy}, Quach~\textit{et al.}~\cite{quach2019learning}, and Guarda~\textit{et al.}~\cite{Guarda2021AdaptiveDL} apply the uniform voxel model in Fig.~\ref{fig:data_strucutre}a to represent voxelized PCG, in which 3D dense convolutions can be directly applied to capture spatial correlations across  voxels within the receptive field. These 3D convolutional layers are often stacked in an autoencoder architecture with quantized latent features at the bottleneck layer compressed using an adaptive arithmetic coder. For any given feature element at the bottleneck, its hyperprior and autoregressive neighbors can be utilized for probability estimation~\cite{Wang2021Lossy}.

{\bf Octree Representation.} As in the uniform voxel representation,  empty voxels, i.e., NOVs, are treated equally as those POVs in computation, leading to unbearable consumption of storage and computation. 
Then, the octree model is revisited since it represents POVs from coarse to fine scales in a progressive manner. Huang~\textit{et al.}~\cite{Huang2020OctSqueezeOE} and Biswas~\textit{et al.}~\cite{NEURIPS2020} adopt the Multi-Layer Perceptron (MLP) to exploit the dependency between parent and child nodes for occupancy probability prediction.

Although the octree model utilizes the occupancy to decompose the input PCG adaptively, its efficiency is largely constrained  because of noticeable bits overhead caused by signaling the deep tree hierarchy incurred by isolated sparse points.

{\bf Hybrid Model.} In addition to solely relying on the parent-child dependency of the octree model, available sibling nodes  can be organized under a uniform voxel representation to use dense 3D CNNs (Convolutional Neural Networks), or under a series of sequential nodes to apply the Transformer, for context modeling. Such hybrid model is extensively studied in~\cite{Que2021VoxelContextNetAO,Fu2022OctAttentionOL} with noticeable  compression performance improvement. 

{\bf Point-wise Representation.} Point-wise solutions directly compress raw points of input point cloud without voxelization. They typically apply the  PointNet++~\cite{Qi2017PointNetDH} or other MLP-based autoencoders. As reported by Gao {\it et al.}~\cite{Gao2021PCGCGraphSampling}, point-wise approaches not only perform poorly at high bit rates, but also have difficulties to generalize themselves to large-scale point clouds, although they have a low complexity requirement.

As will be shown later, the proposed SparsePCGC can effectively solve problems in existing solutions with better coding efficiency, lower complexity, and robust generalization. 
To help the audience understand the proposed framework quickly, we will review the processing of sparse tensor next.

\subsection{Sparse Tensor Processing} \label{sec:STP}

\subsubsection{Sparse Tensor}

The number of POVs or MP-POVs is just a small percentage of total voxels in a volumetric PCG, exhibiting very sparse distribution in a 3D space. Thus, we choose to use the {\it Sparse Tensor} in Fig.~\ref{fig:data_strucutre}c to represent a voxelized PCG, with which a hash map is used to index the geometric coordinates (and associated attributes if applicable) of those MP-POVs {and an auxiliary data structure is applied to manage their connections efficiently~\cite{choy20194d}.} As such, we greatly reduce the time complexity by just performing the computations on sparse MP-POVs to aggregate and embed information from available spatial neighbors.

By contrast, the octree model is limited by its parent-child hierarchy, which is difficult to represent spatial neighbors unless we explicitly construct such local spatial connections through the use of CNN or Transformer as in~\cite{Que2021VoxelContextNetAO,Fu2022OctAttentionOL}; {while the uniform voxel model processes all voxels regardless of their occupancy status, which is redundant obviously.}

\subsubsection{Sparse Convolution} 

3D  sparse convolution can be applied on sparse tensors efficiently. It is similar to commonly-used 3D dense convolution but only convolves using valid {MP-POVs}, which makes full use of the sparse characteristics of point clouds. 

There are a variety of implementations of sparse convolution, e.g.,  Sparse 3D convolution~\cite{BMVC2015_150}, Sparse Blocks Network~\cite{ren2018sbnet}, SpConv~\cite{Yan2018SECONDSE}, Submanifold Sparse ConvNet (Convolutional Network)~\cite{SubmanifoldSparseConvNet}, and 4D Spatiotemporal ConvNets (Minkowski CNN)~\cite{choy20194d}. This work chooses the 4D Spatiotemporal ConvNets compliant MinkowskiEngine~\cite{choy20194d} to demonstrate the SparsePCGC due to its wide support of different operators and platforms. Other implementations can be used as well, for instance,  recently-released TorchSparse\cite{tang2022torchsparse} that offers 1.6$\times$ speedup of the MinkowskiEngine~\cite{choy20194d} can be a promising option for much faster prototype.

A sparse tensor can be formulated using a set of coordinates $\vec{C}=\{(x_i, y_i, z_i)\}_i$ and associated features $\vec{\bf F}=\{\vec{f}_i\}_i$. Thus the sparse convolution is formulated as :
\begin{equation}
\vec{f}_{u}^{out} = \sum\nolimits_{k \in \mathbb{N}^{3}(u, \vec{C}^{in})} W_{k} \vec{f}_{u+k}^{in} 
\quad\text{for}\quad
u \in \vec{C}^{out},
\end{equation} 
where $\vec{C}^{in}$ and $\vec{C}^{out}$ are input and output coordinates. $\vec{C}^{in}$ and $\vec{C}^{out}$ are the same if the resolution is kept. 
$\vec{f}_{u}^{in}$ and $\vec{f}_{u}^{out}$ are input and output feature vectors at coordinate $u$ (a.k.a., ($x_u$, $y_u$, $z_u$)), respectively. 
$\mathbb{N}^{3}(u, \vec{C}^{in}) = \{k|u+k \in \vec{C}^{in}, k\in \mathbb{N}^{3}\}$ defines a 3D convolutional kernel centered at $u$ with offset $k$ in $\vec{C}^{in}$. By setting different $k$, we can easily retrieve neighboring POVs or MP-POVs within a predefined 3D receptive field, e.g., $k\times k \times k$ or $k^3$, for information aggregation. $k$ is set to 3 or 5 for lightweight computation. $W_i$ is kernel weights. 

Clearly, both space and time complexity is greatly reduced when using sparse convolution instead of regular dense convolutions.  More details about the implementation of sparse convolution used in this work can be found in~\cite{choy20194d,MinkowskiEngine}. Just as a quantitative reference, our GPU-based SparsePCGC requires less encoding and decoding runtime than the CPU-based G-PCC anchor, consuming just a few percentage of those uniform voxel based methods~\cite{Nguyen2021LosslessCO, Nguyen2021MultiscaleDC} (see Table~\ref{table:gpcc}). 
In the meantime, our method requires a small amount of storage space due to the model sharing strategy across different scales (see Sec.~\ref{sec:perf_discussion}).

\subsubsection{Notation}
Sparse convolution (SConv) or  transposed sparse convolution (TSConv) can be formatted using ``$K$, $C$, $S$'', where $K$ = $k\times k\times k$ ($k^3$) is the receptive field of 3D convolution, $C$ describes the number of channels, and $S$ can be $s\times s \times s\downarrow$ ($s^3\downarrow$)  for downscaling  or $s\times s \times s\uparrow$ ($s^3\uparrow$) for upscaling. Note that downscaling (upscaling) operator is associated with the SConv (TSConv) accordingly. Except for voxel sampling layer in Fig.~\ref{fig:basic_modules_VSL}, resolution scaling is not required for most cases with $s=1$, we can then simplify the notation to ``SConv $k^3$, $C$'' by omitting the $S$.

%% file: 3_method_part1.tex
\begin{figure}[t]
\centering
	\subfloat[]{\includegraphics[width=2.5in]{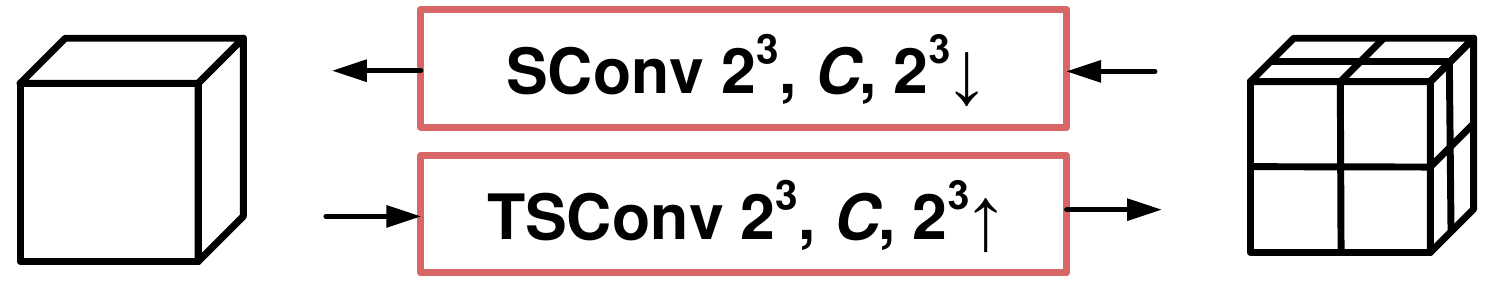}\label{fig:basic_modules_VSL}}\\
	\subfloat[]{\includegraphics[width=2.4in]{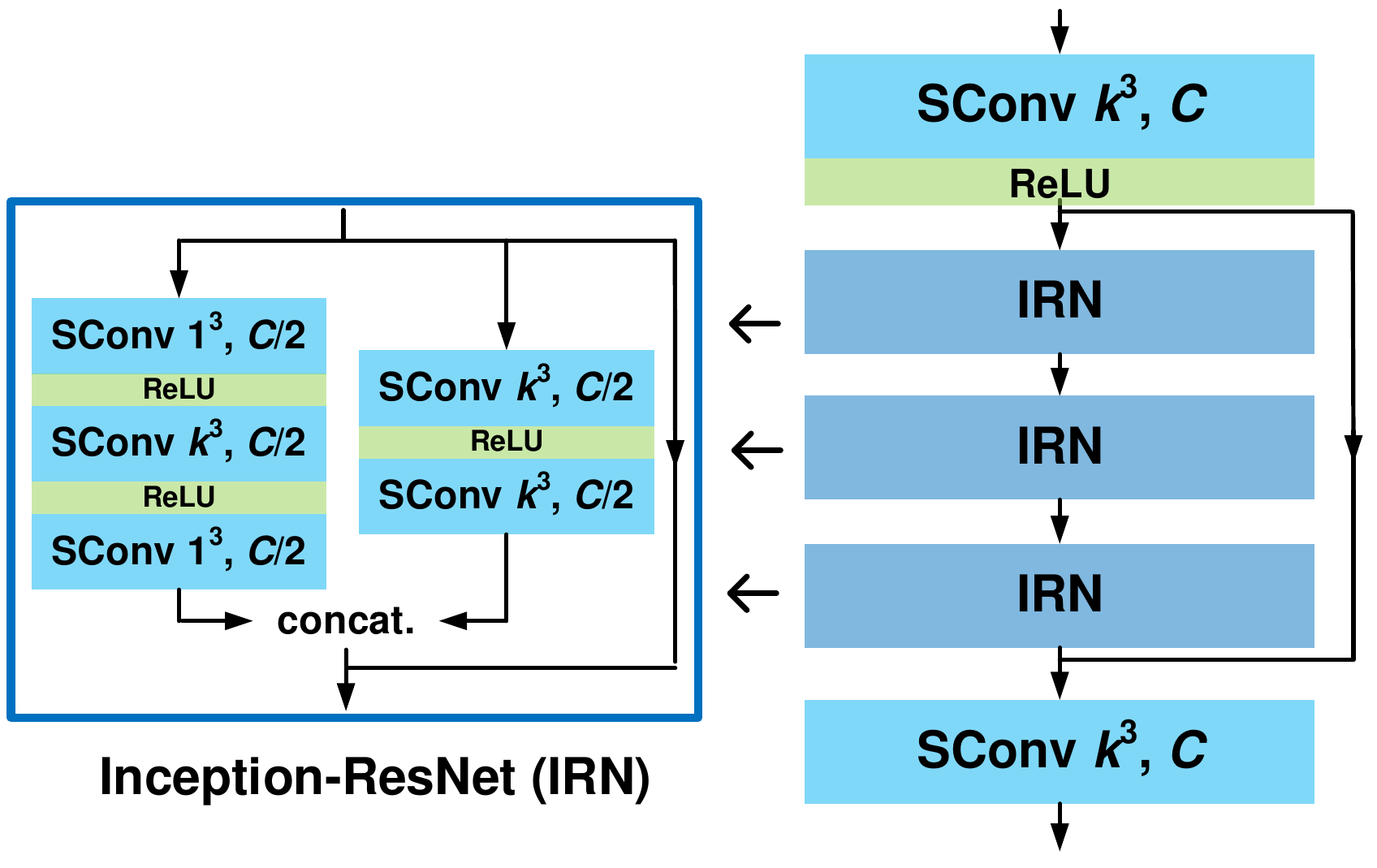}\label{fig:basic_modules_DFA}}
	\subfloat[]{\includegraphics[width=0.7in]{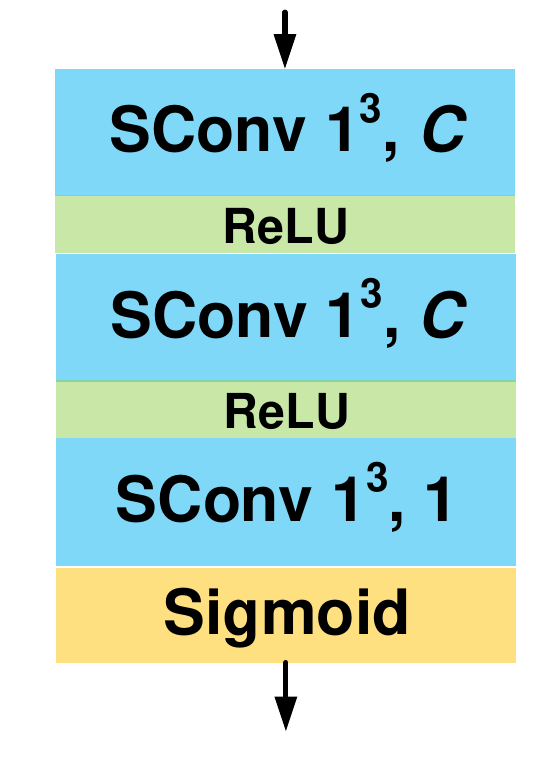}\label{fig:basic_modules_OOL}}

	\caption{
	{\bf Illustration of {Basic SparseCNN Modules.}} (a) Voxel Sampling Layer (VSL) applied to upscale or downscale voxels dyadically through the use of ``SConv $2^3, C, 2^3\downarrow$'' or ``TSConv $2^3, C, 2^3\uparrow$''; (b) Deep Feature Aggregation (DFA) used to characterize and embed information from spatial neighbors within the receptive field; (c) Occupancy Output Layer (OOL) for probability or offset derivation.}
	\label{fig:basic_modules}
\end{figure}

\section{Multiscale Sparse Tensor-based PCGC}\label{sec:method}

This section first pictures the overall system design and then offers technical details of the proposed SparsePCGC.

\subsection{Framework}
A general framework of multiscale sparse tensor-based PCGC is illustrated in Fig.~\ref{fig:overview}. We call it SparsePCGC for short\footnote{Note that we first exemplify the idea using lossless PCGC; then, the proposed architecture can be easily extended to support lossy PCGC. Our preliminary results on lossy PCGC can also be found in~\cite{Wang2020MultiscalePC}.}. 
It includes a pair of encoder and decoder, where the encoder inputs the PCG to generate the compressed bitstream, and the decoder parses the bitstream to reconstruct the original PCG. 

The framework contains a sequence of dyadic voxel sampling, e.g., ${S}$ = $2\times 2\times 2\uparrow$ as the upscaling, or ${S}$= $2\times 2\times 2\downarrow$ as the downscaling. They can be integrated with the SparseCNN blocks for better information embedding during the resampling. 
Such a progressive scaling mechanism enables the multiscale representation so that we can support resolution scalability easily. 
Since we enforce the identical processing between two adjacent scales, the overall SparsePCGC can be comprehensively explained through a two-scale example from the ($i$-1)-th scale to the $i$-th scale as  in Fig.~\ref{fig:overview}.

As seen, each POV from the preceding scale will be sub-divided into eight MP-POVs (e.g., expanding from Fig.~\ref{fig:data_strucutre}d to~\ref{fig:data_strucutre}e), of which some may be POVs and the rest are NOVs. 
The ground truth is known in the encoder but unknown in the decoder.  At each scale, we need to compress the {occupancy status} of each MP-POV in the encoder, and reversely reconstruct POVs from upscaled MP-POVs in the decoder by interpreting compressed syntax. 
Thus, the same cross-scale context model is shared between encoder and decoder, which is facilitated by the SOPA using decoded POVs in the preceding scale to generate occupancy probability $p_{\text{MP-POV}}$ of the MP-POV in upscaled tensor for arithmetic coding.
 
Next, we first brief basic SparseCNN blocks used for SOPA computation in SparsePCGC.

\subsection{Basic SparseCNN Blocks}

Convolutional and (nonlinear) activation layers are generally stacked to form SparseCNN blocks, including  voxel sampling layer (VSL), deep feature aggregation (DFA), and occupancy output layer (OOL) shown in Fig.~\ref{fig:basic_modules}. All computations are performed using sparsely distributed POVs or MP-POVs in a sparse tensor.

\begin{figure}[t]
\centering
\includegraphics[width=3.4in]{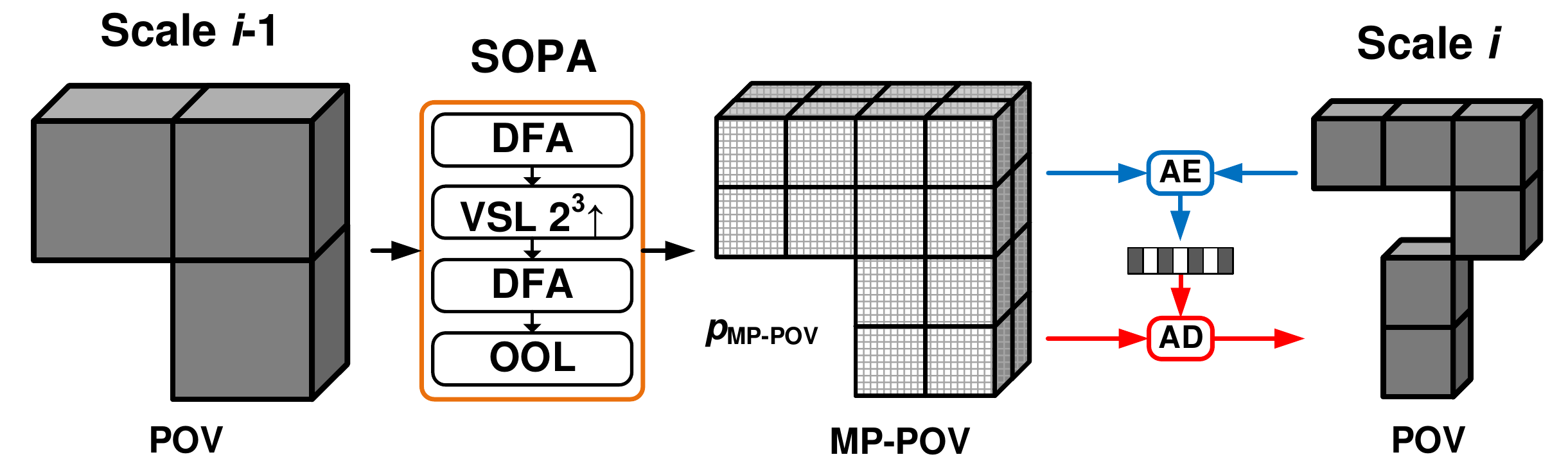}
\caption{{\bf One-Stage SOPA.} Each decoded POV is processed by the SOPA engine to generate eight MP-POVs and associated $p_{\text{MP-POV}}$s. The encoder uses $p_{\text{MP-POV}}$s to encode ground truth into the bitstream while the decoder determines true POVs and NOVs by decoding the bitstream with $p_{\text{MP-POV}}$s.}
\label{fig:single_stage_SOPA}
\end{figure}

{\bf Voxel Sampling Layer (VSL).}  We embed voxel upscaling in SOPA and voxel downscaling in SLNE\footnote{ If we do not include the SLNE in the encoder, simple dyadic voxel downscaling is used.} to execute the cross-scale message passing. 
For the downscaling in SLNE illustrated in Fig.~\ref{fig:SLNE}, we apply the ``SConv $2^3$, $C$, $2^3\downarrow$'' to merge eight spatially-connected voxels (e.g., $2\times2\times2$)  into a single voxel, as shown in Fig.~\ref{fig:basic_modules_VSL} which is noted as ``VSL $2^3\downarrow$'' for short. 
For the upscaling in SOPA depicted in Fig.~\ref{fig:single_stage_SOPA} and Fig.~\ref{fig:multi-stage-SOPA}, we apply the ``TSConv $2^3$, $C$, $2^3\uparrow$''  to sub-divide a voxel into eight sub-voxels (or child nodes), i.e.,  VSL $2^3\uparrow$. $C$ is an adjustable model parameter standing for the number of channels.

{\bf Deep Feature Aggregation (DFA).}
For most convolutional layers without resolution scaling, we typically apply the ``SConv $k^3$, $C$'' to aggregate features of neighboring voxels at the same scale. We set $k$ to 3 for dense object point clouds and 5 for sparse LiDAR point clouds because a slightly larger convolutional kernel is more beneficial for much sparser LiDAR PCGs to include sufficient neighbors for information embedding. Increasing $k$ can improve the compression efficiency at the increase of both space and time complexity. The number of channels ($C$) is  set to 32. We use a deep Inception ResNet (IRN) block that is comprised of three basic IRN units~\cite{Szegedy2017Inceptionv4IA} to characterize the spatial dependency across voxel neighbors.  The IRN is also used in our earlier works, demonstrating convincing efficiency and affordable complexity~\cite{Wang2021Lossy, Wang2020MultiscalePC}.

{\textbf{Occupancy Output Layer (OOL).}}
As for {SOPA models} detailed subsequently, the last occupancy output layer (OOL) is applied, which generally consists of three convolutional layers and a Sigmoid activation layer to derive the occupancy probability $p$ in the range of [0,1] for dense object PCG, as illustrated in Fig.~\ref{fig:basic_modules_OOL}. On the other hand, for the compression of sparse LiDAR point clouds, the OOL embedded in SOPA (Position) model removes the Sigmoid layer and expands the number of output channels of the last convolutional layer from 1 in Fig.~\ref{fig:basic_modules_OOL} to 3 in Fig.~\ref{fig:lossy_offset} to derive three coordinate offsets directly.

As will be unfolded later, the aforementioned VSL, DFA, and OOL blocks will be deliberately stacked to form the SOPA, SOPA (Position), and SLNE modules used in SparsePCGC.

\subsection{Lossless SOPA: Exploiting Spatial Priors For More Compact Representation} \label{ssec:sparse_prior}

The compression performance of SparsePCGC is mainly determined by the efficiency of underlying SOPA engine for context modeling. Even for SLNE, it is also utilized to improve the SOPA for a more accurate probability approximation. 
We start our discussion from a toy baseline.

\subsubsection{A Toy Baseline Using One-Stage SOPA Only} 

This example only allows dyadic voxel downscaling (i.e., no SLNEs) in encoder, and ultimately relies on the One-Stage SOPA for MP-POV probability (i.e., $p_{\text{MP-POV}}$) estimation. 
As shown in  Fig.~\ref{fig:single_stage_SOPA}, it inputs each POV from preceding scale to generate $p_{\text{MP-POV}}$s of corresponding upscaled 8 MP-POVs.
Such one-stage SOPA applies the ``TSConv $2^3$, $C$, $2^3\uparrow$'' in VSL, i.e., VSL $2^3\uparrow$, for resolution upscaling. 
In fact, we cannot learn sufficient information if we use the geometry information of preceding lower-scale POVs only for $p_{\text{MP-POV}}$ estimation in such a one-shot manner, yielding inferior coding efficiency to the G-PCC anchor (i.e., average 5.5\% compression ratio increase in Table~\ref{table:models} for lossless coding).

\begin{figure}[t]
\centering
\subfloat[]{\includegraphics[width=0.5in]{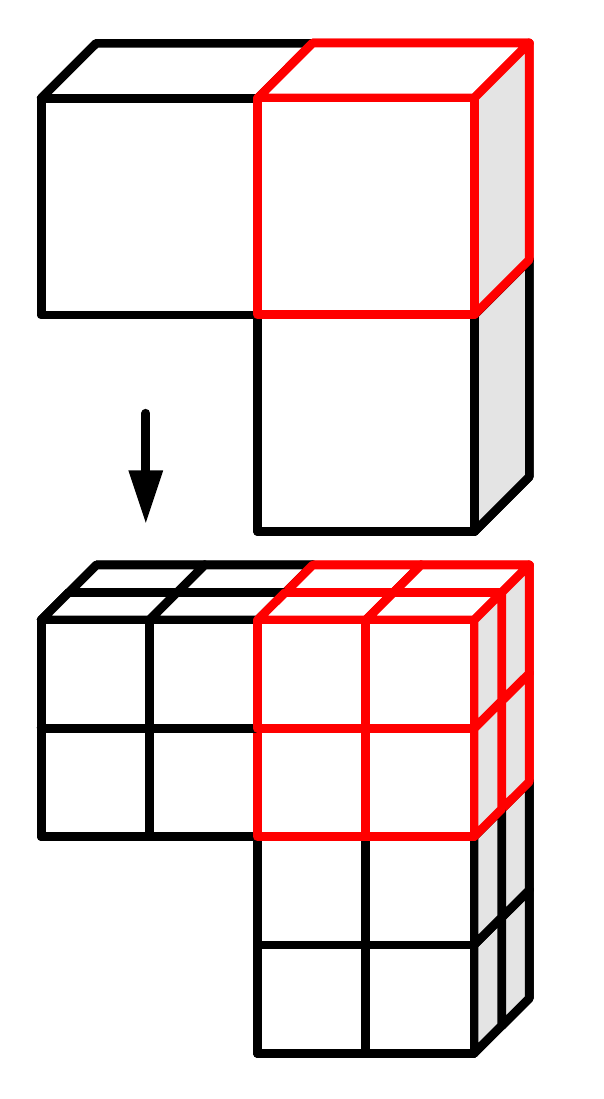}}
\subfloat[]{\includegraphics[width=2.8in]{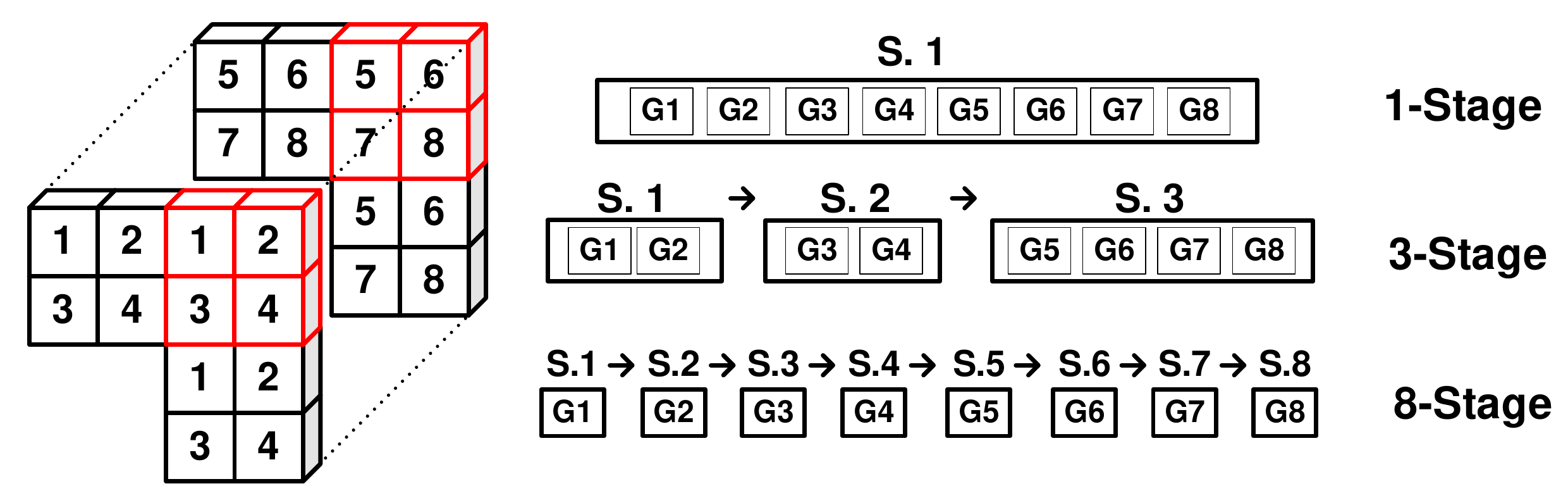}\label{sfig:grouping}}
\caption{{\bf Grouping Arrangement for Multi-Stage Processing.} (a) dyadic resolution upscaling: one POV is upsampled to eight MP-POVs (or child nodes as illustrated by octree expansion); (b)  1/3/8-Stage examples by grouping eight labeled MP-POVs accordingly. Numbers with symbol $G$ correspond to group labels, i.e., elements marked with ``$i$'' belong to the $i$-th group, e.g., G$i$, $i=$1, 2, $\ldots$, 8. Elements in different groups can be categorized for parallel processing at each stage (e.g., S.$j$).}
\label{fig:partition}
\end{figure}

\begin{figure*}[t]
\centering
\includegraphics[width=7.0in]{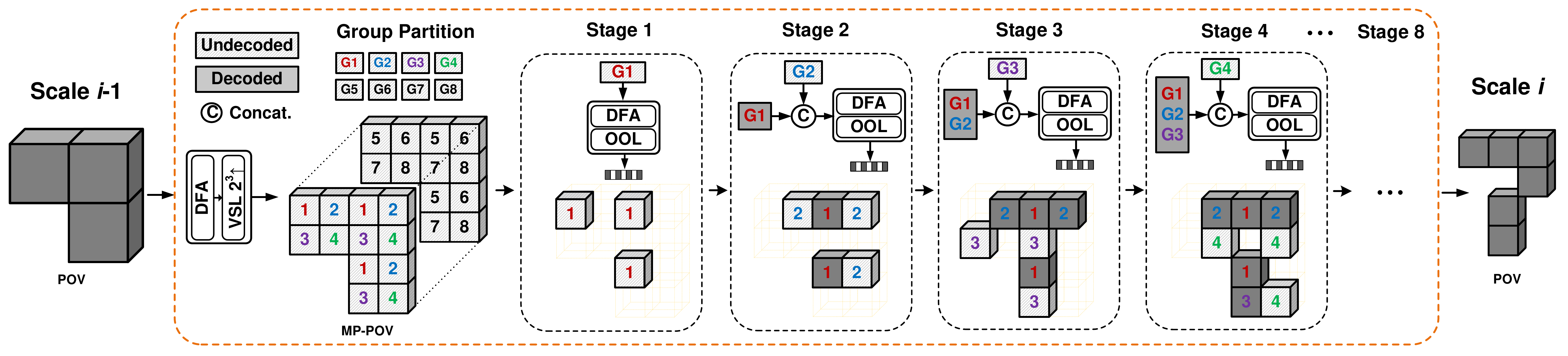}
\caption{{\bf Multi-Stage SOPA.} 8-Stage SOPA is exemplified for cross-scale context modeling. At Stage $n$, MP-POVs in G$n$ are processed together with the POVs and their associated features that are determined in preceding stages, e.g., $n-1$, $n-2$, $\ldots$, for occupancy probability estimation. Note that the occupancy probability is used in both encoder and decoder.}
\label{fig:multi-stage-SOPA}
\end{figure*}

\subsubsection{Multi-Stage SOPA}

Previous One-Stage SOPA assumes that  eight MP-POVs upscaled from a POV of the last scale are independent of each other. However, these MP-POVs are closely correlated because they are generated from the same POV and are present in close proximity. Hence, this section develops the Multi-Stage SOPA to progressively estimate the $p_{\text{MP-POV}}$s by exploiting the correlations of preceding lower-scale POVs and same-scale causal neighbors.
The causal neighbors are already processed and serve as the prior knowledge for the process of current elements.
They are also widely used in learning-based image/video coding~\cite{minnen2018joint,liu2019non} for more accurate probability estimation.

{\bf Grouping Arrangement for Multi-Stage Processing.} Hereafter, the Multi-Stage SOPA is mainly about how to arrange MP-POV groups (e.g., elements from G1 to G8 in Fig.~\ref{sfig:grouping}) for inter-group dependency exploration. We have exemplified three typical grouping arrangements in Fig.~\ref{sfig:grouping}, e.g., a One-Stage example by processing all eight groups in parallel at the same stage ``S. 1'' and an 8-Stage example to process element groups from ``S. 1'' to ``S. 8'' stage by stage. Different groups can be arranged to fulfill the multistage computing, and element groups at the same stage are processed concurrently. To best exploit neighborhood correlation at the same scale, (grouped) POVs that have been just determined in preceding stages are utilized as priors to process MP-POVs in succeeding stages. Next, we use the 8-Stage example (a.k.a., eight groups) to explain the proposed Multi-Stage SOPA. The same methodology can be extended to other grouping schemes easily.

In the companion supplemental material, we explain the grouping arrangement for $n$-Stage SOPA, where elements of all upscaled MP-POVs are processed sequentially in a raster scan order. The $n$-Stage SOPA fully relies on the autoregressive context model for probability estimation, imposing strict causal data dependency in decoding, which yields unbearable runtime (see Table~\ref{table:models}).

{\bf 8-Stage SOPA.} 
Given any ``8 MP-POVs'' set upsampled from the corresponding POV of the preceding scale, an intuitive  scheme is to categorize each element into an individual group, e.g., G1, G2, $
\ldots$, G8 as shown in Fig.~\ref{sfig:grouping}.
The same labeling is applied to all ``8 MP-POVs'' sets. Therefore, MP-POVs labeled with ``1'', i.e., G1 MP-POVs, will be processed simultaneously at the first stage (i.e., ``S. 1''), and then MP-POVs labeled with ``2'' (a.k.a., G2 MP-POVs in the second stage ``S. 2''), till we traverse all eight groups.

\begin{figure*}[t]
\centering
\includegraphics[width=6.8in]{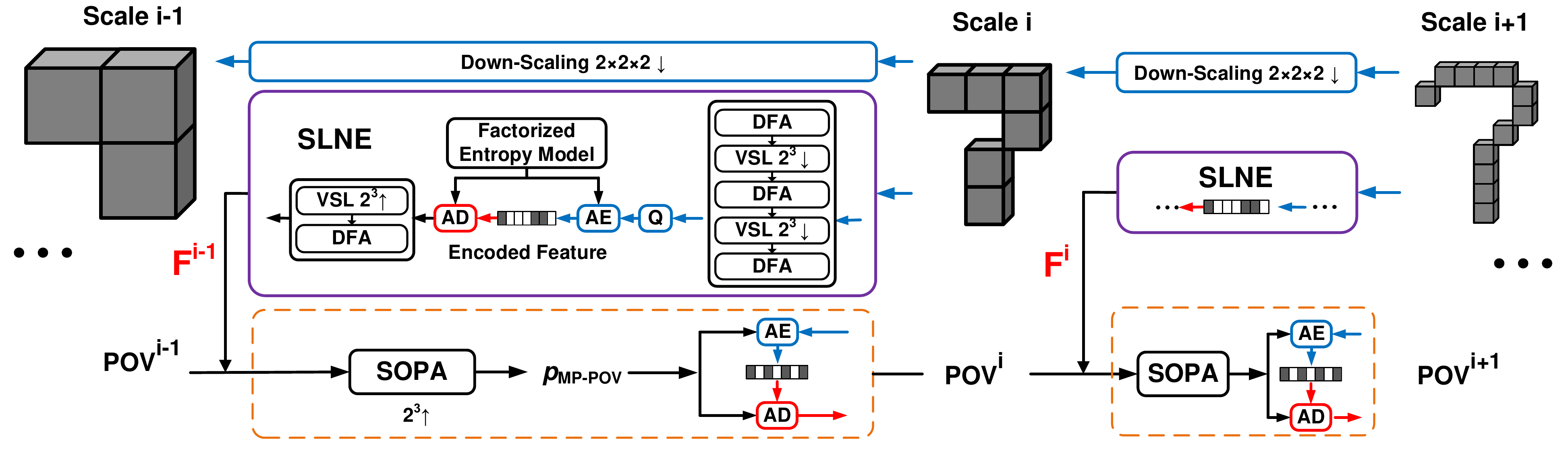}
\caption{{\bf SLNE enhanced One-Stage SOPA.} A two-scale SLNE model interleaves three DFA and two ``VSL $2^3\downarrow$'' blocks  for feature downscaling and embedding; 
correspondingly a pair of ``VSL $2^3\uparrow$'' and DFA block is used for feature upscaling.  Quantization {(Q)} that is widely used in learning-based image/video coding~\cite{liu2019non,balle2018variational} is applied where the uniform noise injection is used in training and rounding is used in inference. The factorized entropy model is applied  for the compression of features~\cite{balle2018variational}. At each scale, the occupancy status and feature attribute of the sparse tensor are separately encoded and multiplexed into the bitstream.}
\label{fig:SLNE}
\end{figure*}

Figure~\ref{fig:multi-stage-SOPA} details the 8-Stage SOPA step by step: 
\begin{enumerate}
    \item All MP-POVs are upscaled from corresponding POVs of preceding lower-scale through the use of stacked DFA and VSL blocks and are partitioned into eight groups from G1 to G8. Multi-Stage SOPA is applied to process grouped elements from the first group G1 to the last group G8, and elements in the same group are processed in parallel.
    \item At the first stage, G1 MP-POVs are processed using stacked DFA and OOL blocks to determine their $p_{\text{MP-POV}}$s for compressing ground truth voxel occupancy information into the bitstream in encoder or parsing bitstream in decoder to identify POVs and NOVs. Note that decoded POVs and associated features (i.e., dark-grey painted ``1'' in Stage 2 of Fig.~\ref{fig:multi-stage-SOPA}) are kept to process MP-POVs in succeeding stages, and the NOVs are pruned immediately, i.e., the upper-left ``1'' in Stage 1 is a NOV which is removed in Stage 2 of Fig.~\ref{fig:multi-stage-SOPA}.
    \item For the second and the remaining stages, computations are the same as ``Stage 1'' but with slightly different input data. As aforementioned, POVs and their features from preceding stages are used as priors to process MP-POVs in succeeding stages for better occupancy probability estimation. 
\end{enumerate}

For either Multi-Stage or One-Stage SOPA, we can approximate the bit rate of MP-POV occupancy status at the $i$-th scale using
\begin{align}
    R_{\text{s}}(i) = \sum\nolimits_j-\log_2(p^{i}_{\text{MP-POV}}(j)), \label{eq:occupancy_status_MPPOV}
\end{align}
where $p^{i}_{\text{MP-POV}}(j)$ is occupancy probability of the $j$-th MP-POV at the $i$-th scale  derived using priors from preceding scale or stages. 

By far, the SOPA model only uses the occupancy status (i.e., 1 or 0) of POVs from preceding scale to compute occupancy probability of current-scale MP-POVs through either a single-stage or a multi-stage manner. Next section will discuss possibilities to learn and embed spatial priors as the feature attribute in encoder to improve the SOPA efficiency.

\subsubsection{SLNE Enhanced  SOPA}\label{subsec:SLNE}

Since we know ground truth labels at each scale in the encoder, we can learn and embed local spatial variations as priors in feature attribute. In this case, for each POV, besides its occupancy status signaling, these local variation features will be compressed and decoded to enhance the capacity of the SOPA engine. This section exemplifies the SLNE-enhanced One-Stage SOPA. Other combinations can be easily extended.

Figure~\ref{fig:SLNE} illustrates the SLNE between the $i$-th and the $(i-1)$-th scales.  Similar to the SOPA model, the same SLNE is applied to any two adjacent scales. As seen, in addition to geometrically downscaling the sparse tensor {\bf POV}$^i$ to the {\bf POV}$^{i-1}$, the SLNE aggregates local neighborhood variations for each POV as its  feature attributes ${\bf F}^{i-1}$. Thus, both occupancy status and feature attributes of each POV in {\bf POV}$^{i-1}$ are compressed into the bitstream, where the rate of occupancy status uses  \eqref{eq:occupancy_status_MPPOV}, and the rate of lossily coded feature attributes is
 \begin{align}
     R_{F}({i-1}) = \sum\nolimits_j -\log_2(p_{{\bf F}^{i-1}}(j)), \label{eq:feature_attribute_rate}
 \end{align}  where $p_{{\bf F}^{i-1}}$ follows the factorized entropy model~\cite{balle2018variational} with its parameters determined in training.
 
From the decoding point of view, both reconstructed occupancy status and features of each POV at the $(i-1)$-th scale are fed into the SOPA engine to derive the occupancy probabilities of MP-POVs at the $i$-th scale. The occupancy reconstruction is achieved by decoding the occupancy status related bitstream using estimated MP-POV probabilities; while the feature reconstruction stacks a DFA block and a VSL block to upscale decoded features by a factor of 2 in three axes accordingly.
 
As illustrated in Fig.~\ref{fig:SLNE}, the SLNE related feature attribute processing, e.g., downscaling, encoding, decoding, and upscaling, are contained within two consecutive scales.  Interleavedly stacking three DFA and two VSL blocks before quantization could well balance the compactness and efficiency of quantized features for optimally estimating the occupancy probability of MP-POVs in the $i$-th scale tensor according to our extensive simulations. Having a pile of one  DFA block and one VSL block to upscale decoded features is to match the geometry resolution of the sparse tensor of preceding lower scale, for instance, the ($i-1$)-th scale in Fig.~\ref{fig:SLNE}.

%% file: 3_method_part2.tex
\begin{figure}[t]
\centering
\subfloat[]{\includegraphics[width=3.4in]{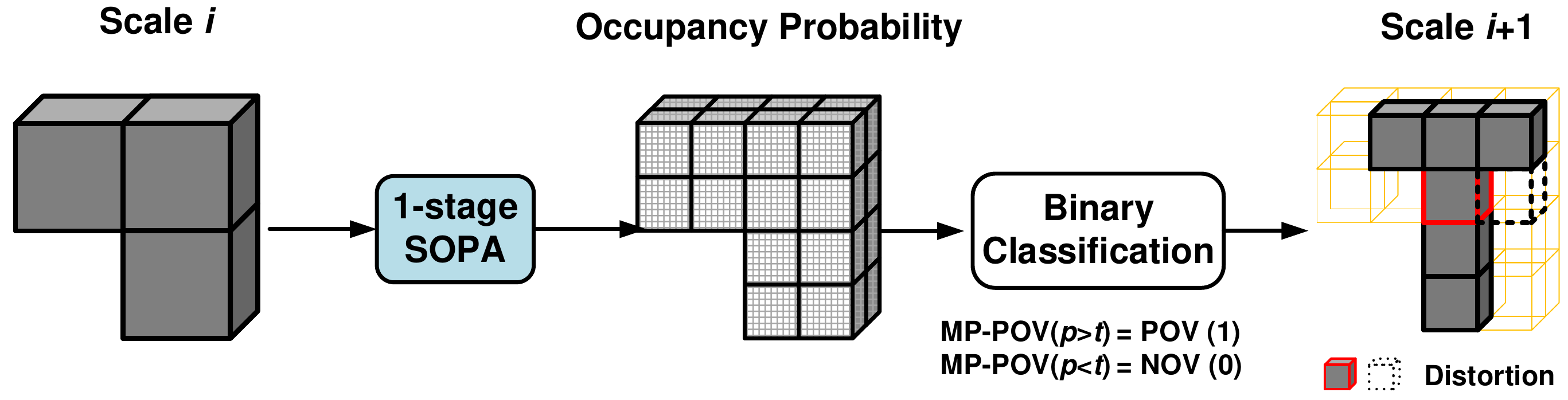}\label{fig:lossy_occupancy}}
\quad\quad
\subfloat[]{\includegraphics[width=3.4in]{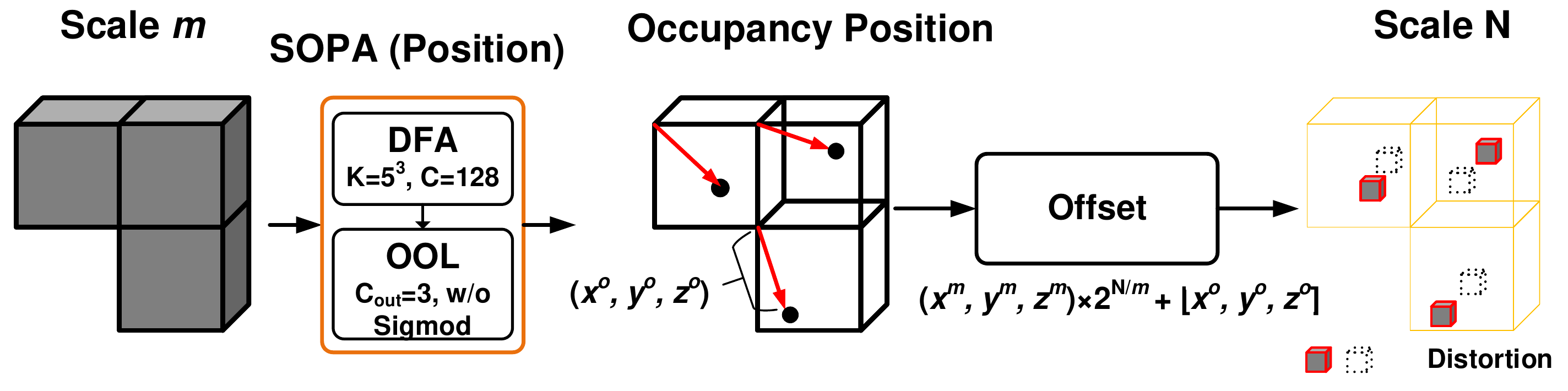}\label{fig:lossy_offset}}
\caption{{\bf Lossy SOPA.} (a) Probability thresholding for dense object point clouds; (b) Position offset adjustment (POA) for sparse LiDAR point clouds.}
\end{figure}

\section{Unified Lossless and Lossy Compression}\label{sec:method_lossy}

Past sections detail the SOPA variants in lossless mode. Here, we show that it can be extended to support lossy compression under a unified framework. A slight modification is that we apply different strategies in lossy SOPA engine for dense and sparse point clouds as they present  very diverse characteristics, including point density, bit precision, etc.

\subsection{Lossy SOPA}

\subsubsection{Probability Thresholding for Dense Point Clouds}\label{sec:method_lossy_prob}

Recalling that we use the OOL block in SOPA shown in Fig.~\ref{fig:single_stage_SOPA} to derive the occupancy probability of each MP-POV in lossless mode, the lossy mode then simply applies a probability threshold $t$ to classify and determine the binary occupancy status. For instance, as illustrated in Fig.~\ref{fig:lossy_occupancy},
if $p_{\text{MP-POV}}>t$, the \text{MP-POV} is a \text{POV}; otherwise it is a \text{NOV}. $t$ is set adaptively according to the number of POVs at each scale, which is the same as our previous works~\cite{Wang2021Lossy, Wang2020MultiscalePC}. Apparently, distortion is inevitable due to false classification.

As for the Multi-Stage SOPA model, the probability estimation in the succeeding stage relies on the outcome of preceding stages, implying that a false classification at earlier stages may accumulate errors stage by stage and lead to unexpected distortions. Thus, we choose to use One-Stage SOPA for lossy compression in this work.

\subsubsection{Position Offset Adjustment for Sparse Point Clouds}

The proposed probability thresholding-based lossy SOPA works well for the compression of dense point clouds. However, it fails in LiDAR point clouds because of  the  extremely sparse distribution of points. Particularly at higher scales, occupancy estimation of the MP-POV is inefficient.

To solve this issue, we replace the occupancy {\it probability approximation} in the native SOPA model by occupancy {\it position adjustment}. For clarification, it is referred to as the ``SOPA (Position)'' in Fig.~\ref{fig:lossy_offset}.
As seen, it consists of a DFA  block and a modified  OOL block to directly estimate the coordinate offsets. Compared with the SOPA model in Fig.~\ref{fig:lossy_occupancy}, the VSL block used for resolution scaling is removed. $k = 5$ and $C = 128$ are used to capture sufficient information from sparse LiDAR points.  In the OOL block, the Sigmoid function is removed and the output layer is set with 3 channels to produce the position offsets for adjustment.

For a sparse tensor at the $m$-th scale, we upscale it to the $N$-th scale in one-shot ($ N > m$) as in Fig.~\ref{fig:lossy_offset}. For a given POV at $(x^{m}, y^{m}, z^{m})$ in the $m$-th scale, its corresponding POV at the $N$-th scale locates at
\begin{equation}
    (\hat{x}^{N}, \hat{y}^{N}, \hat{z}^{N}) = (x^{m}, y^{m}, z^{m})\times 2^{\frac{N}{m}} + \lfloor x^{o}, y^{o}, z^{o} \rceil,
\end{equation} with $(x^{o}, y^{o}, z^{o})$ as the position offset estimated from the proposed SOPA (Position). 

Although the position adjustment expands the resolution, it does not increase the number of POVs, which well reflects the distribution characteristics of LiDAR sequences.
Similar position offset adjustment methodology is also used in~\cite{Que2021VoxelContextNetAO}, where it is called ``Coordinates Refinement Module {(CRM)}''. Note that either position offset adjustment or CRM belongs to the post-processing technique, which is optional for compression performance evaluation~\cite{Fu2022OctAttentionOL}.

\begin{figure}[t]
\centering
\includegraphics[width=3.5in]{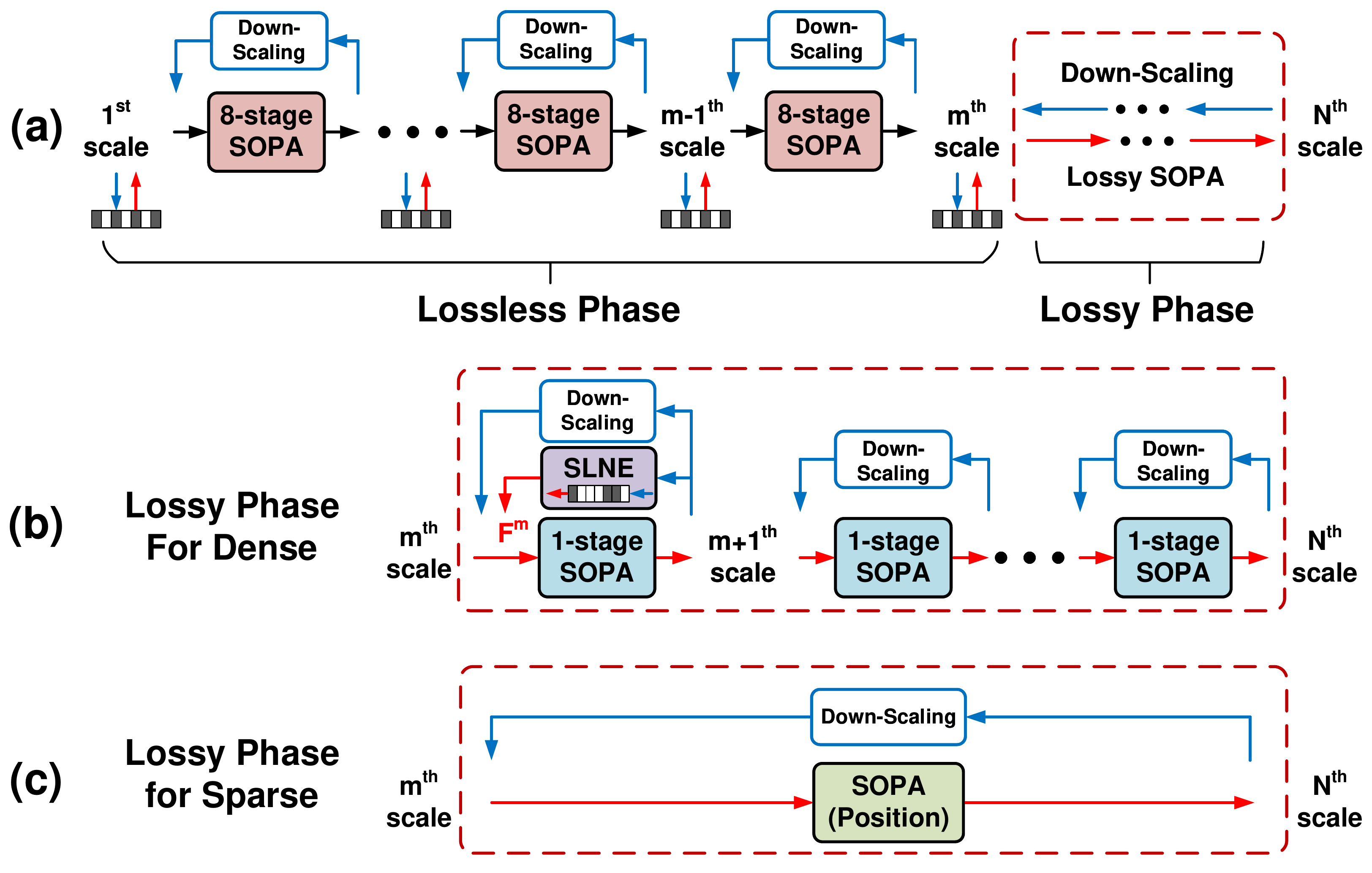}\label{fig:overview_lossy}\\
\caption{{\bf Lossy SparsePCGC.} (a) General architecture with $m$ scales coded losslessly and $(N-m)$ scales coded in lossy mode;  (b) lossy SOPA for dense object point clouds; (b) SOPA (Position) for sparse LiDAR point clouds; }
\label{fig:lossy_SparsePCGC}
\end{figure}

\subsection{Unified SparsePCGC Framework}

Previous discussions focus on the development of models using two adjacent scales in either lossy or lossless mode.
In practice, the SparsePCGC stacks a sequence of two-scale models to  process dense or sparse PCG input, in both lossy and lossless modes. 

In lossless mode, the solution is fairly straightforward: we apply lossless SOPA models to compress both dense and sparse PCGs across all scales. The compression ratio against standardized G-PCC anchor is used to evaluate the performance quantitatively. Comparisons with other learning-based solutions are also provided.

As for lossy mode illustrated in Fig.~\ref{fig:lossy_SparsePCGC},  the SparsePCGC devises the lossless coding mode from the first scale at the lowest resolution to the $m$-th scale, and the lossy mode to encode the remaining scales. Here we assume the $N$ scale in total.  We adapt $m$ to best balance the rate and distortion for lossy compression.
As reported in~\cite{Wang2020MultiscalePC}, applying lossy coded PCG for all scales would produce severe degradation of reconstruction quality due to scale-by-scale refinement, but could not bring noticeable rate reduction.  
 
Note that the highest scale $N$ is typically determined by the geometry precision of input point clouds. For example, $N$ = 10 or 11 for dense 8iVFB~\cite{8i20178i} and Owlii~\cite{xu2017owlii}, and $N$ = 18 for KITTI and Ford sequences.

\section{Training}
This section details the datasets, loss functions, and strategies used to train the proposed SparsePCGC.

\subsection{Datasets} \label{sec:training_data}

Large-scale, popular, and publicly accessible ShapeNet~\cite{Chang2015ShapeNetAI} and KITTI~\cite{Behley2019SemanticKITTIAD} are used to generate respective training datasets for dense object and sparse LiDAR point clouds. 
\begin{itemize}
\item  {\bf ShapeNet}~\cite{Chang2015ShapeNetAI} is one of the most popular CAD model dataset for 3D objects. Its subset ShapeNetCore contains 55 common object categories with about 51,300 unique 3D mesh models. We densely sample points on raw meshes to generate point clouds, and then randomly rotated and quantized them with 8-bit geometry precision at each dimension. 
\item {\bf KITTI} (SemanticKITTI)~\cite{Behley2019SemanticKITTIAD} is a large-scale LiDAR dataset used for semantic scene understanding. It contains 22 sequences, a total of 43,552 scans (or frames) of outdoor scenes collected using the Velodyne HDL-64E LiDAR sensor. There are around 120k points on average per frame.
These raw floating-point coordinates are quantized to millimeter scale (e.g., unit precision at 1mm), requiring 18-bit geometry precision for storage.
\end{itemize}

\textbf{Data Augmentation.}
In practice, the characteristics of point clouds vary greatly from one to another. Among them, geometry precision is one of the most important factors, because it has a significant impact on the spatial variance, density, and thus the efficiency of compression methods.

It is impractical to assume a fixed geometry precision for raw input, and it is also difficult to set a constant number of scales. Recalling that our solution applies the resolution scaling for multiscale representation, we train SOPA models  for two consecutive scales and then apply them for inference across all scales.

In order to generalize the trained models to an excessive amount of dynamic content, we first downscale the point clouds by a random scaling factor in $(0.5,1)$ and then perform dyadic voxel downscaling upon them with the scaling factors at $\{1,\frac{1}{2},\frac{1}{4}, \frac{1}{8}, \cdots\}$ to obtain ground truth labels at multiple scales. 
These ground truth labels are then used to supervise the training of SOPA and SLNE models.

{\bf Remark.} We specifically use ShapeNet trained model to compress dense object point clouds (e.g., 8iVFB, Owlli), and apply KITTI tuned model for sparse LiDAR point clouds like Ford sequences, when quantitatively measuring the compression efficiency against the  G-PCC anchor. This is because very diverse point density and bit precision properties are presented for dense object and sparse LiDAR content, making it inefficient to use a single model for compression. Note that the test sequences exhibit very different content distributions from that of the training set, evidencing the robust generalization of the proposed SparsePCGC (see Table~\ref{table:gpcc}). 

On the other hand, in order to ensure the fair comparison with other learning-based approaches like VoxelDNN~\cite{Nguyen2021LosslessCO}, NNOC~\cite{Kaya2021NeuralNM}, VCN~\cite{Que2021VoxelContextNetAO}, and OctAttention~\cite{Fu2022OctAttentionOL}, we have finetuned our model using  their training sets for performance evaluation and enforced the test using a comparative low-end hardware platform (e.g., RTX 2080) for complexity measurement.
We have also faithfully reproduced the best results of these baselines by either executing publicly-accessible source codes~\cite{Fu2022OctAttentionOL,Kaya2021NeuralNM}, or consulting authors for the latest  data~\cite{Que2021VoxelContextNetAO}, or directly citing numbers from published paper~\cite{Nguyen2021LosslessCO,Nguyen2021MultiscaleDC}.

\subsection{Loss Functions}
Basically, the SOPA model estimates the probability of occupancy status $p_{\text{MP-POV}}$ for a given MP-POV at a specific scale, which is then used to code the binary occupancy symbol of this MP-POV in lossless mode, or to perform the  binary classification in lossy compression mode.

In general, the Binary Cross-Entropy (BCE) between estimated occupancy probability and actual occupancy symbol is used in training, i.e.,
\begin{equation}
\begin{split}
& L_{\text{BCE}} = \\ 
& \sum\nolimits_{j}-\big(o(j)\log_{2}(p(j)) + (1-o(j))\log_{2}(1 - p(j))\big),  \label{eq:BCE}
\end{split}
\end{equation} 
where  $o(j)$ represents the actual occupancy symbol (e.g., 1 for POV and 0 for NOV), $p = p_{\text{MP-POV}}$ is the probability that the $j$-th MP-POV is POV, and thus (1-$p$) is the probability of being the NOV.

Note that the SOPA model can be improved by including encoder-side SLNE model. Thus, the loss function for SLNE enhanced SOPA model is formulated as the combination of BCE loss in \eqref{eq:BCE} and rate consumption of feature attributes: 
\begin{align}
    L_{\text{comb}} = L_{\text{BCE}} + R_F, \label{eq:slne_enh_sopa}
\end{align} with $R_F$ defined in \eqref{eq:feature_attribute_rate}.

For sparse LiDAR point clouds using SOPA (Position) model, we treat it slightly different. Given that the SOPA (Position) model directly outputs the occupancy position offset, we measure the Mean Square Error (MSE) between the coordinates of adjusted points and true points in upscaled sparse tensor, i.e.,
\begin{equation}
   L_{\text{MSE}} = \frac{1}{J}\sum\nolimits_{j} \big((x^{N}, y^{N}, z^{N})(j) - (\hat{x}^{N}, \hat{y}^{N}, \hat{z}^{N})(j)\big)^{2},
\end{equation} having $J$ points in total.

\begin{figure}[t]
	\centering
	\subfloat[]{\includegraphics[width=1.7in]{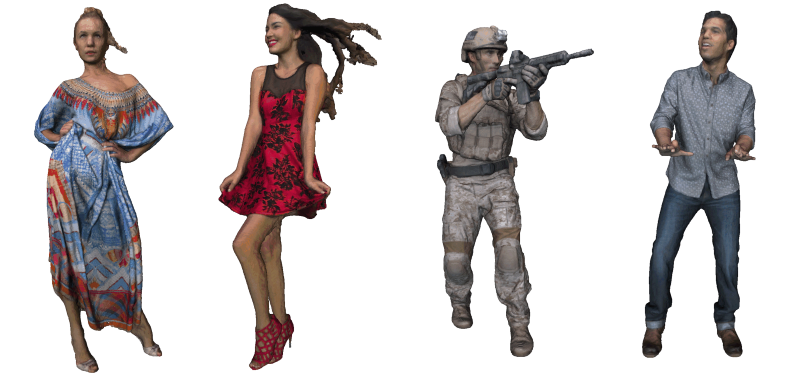}\label{fig:ablation_vis_8iVFB}}
	\subfloat[]{\includegraphics[width=1.8in]{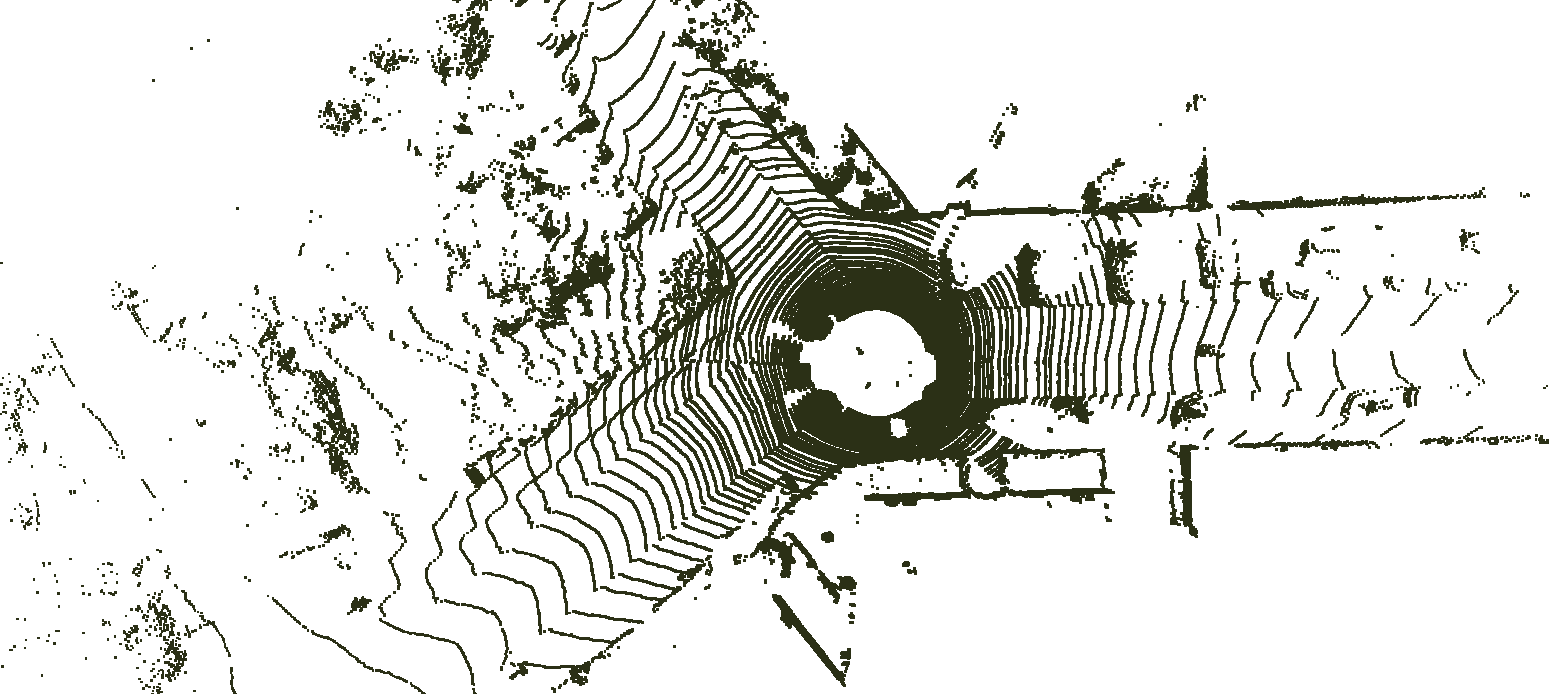}\label{fig:ablation_vis_kitti}}
	\caption{{\bf Point Cloud Examples.} (a) Dense object point clouds from 8iVFB; (b) Sparse LiDAR point clouds from KITTI.}
	\label{fig:various_data}
\end{figure}

\subsection{Training Strategies}
Overall, the training of SOPA models is very straightforward and easy, while jointly optimizing combined loss functions in SLNE enhanced  One-Stage SOPA model is slightly difficult. Therefore we propose to set the weight of $R_{F}$ to 0 in \eqref{eq:slne_enh_sopa} at first, and then gradually increase it to 1 for robust and fast convergence. For SLNE enhanced Multi-Stage SOPA, only the SOPA models are updated, while the SLNE model is directly borrowed from the SLNE enhanced One-Stage model and its parameters are fixed in training.

The learning rate decreases from $0.0008$ to $0.00002$ in the training. The training iteration executes around $30$ epochs with a batch size of 8. Adam is used as the optimizer.

%% file: 4_exp_part1.tex
\input{table/table_multistage}

\section{Experimental Studies}
\label{sec:exp}
\subsection{Test Setup}
{\bf Datasets.} The MPEG PCC datasets~\cite{mpegdataset} that are recommended and used by standardization committee are tested for performance evaluation as illustrated in Fig.~\ref{fig:various_data}.

\textit{Dense Point Clouds.}
We choose several static samples from the Category 1 of MPEG PCC dataset, including longdress\_vox10\_1300, redandblack\_vox10\_1550, soldier\_vox10\_0690, loot\_vox10\_1200, queen\_0200, basketball\_player\_vox11\_0200, and dancer\_vox11\_0001. They are densely sampled but have different geometry precision and characteristics to represent diverse objects. Later, both 8iVFB and MVUB sequences are used when performing comparative studies with other learning-based approaches.

\textit{Sparse Point Clouds.}
We use KITTI and Ford samples as typical sparse LiDAR point clouds. 
For KITTI content, we use 0-11 sequences for training (see Sec.~\ref{sec:training_data}) and 12-21 sequences for testing\footnote{Such split of training and testing sequences was commonly used in~\cite{Huang2020OctSqueezeOE,Que2021VoxelContextNetAO,Fu2022OctAttentionOL}.}. The raw floating-point coordinates are quantized to 1mm (18 bits required) and 2cm (14 bits required) precision.

The Ford dataset is used to test dynamically-acquired point clouds as the Category 3 samples in MPEG PCC datasets. It contains 3 sequences (Ford\_01, Ford\_02, and Ford\_03), and each sequence has 1500 frames and averaged 100k points per frame. We use all 3 sequences for testing. 
The original Ford dataset is already quantized to 1mm (18 bits) precision. Here, we further quantize it to 2cm (14 bits required) precision for an additional test.

More details regarding the MPEG PCC datasets can be found in~\cite{mpegdataset}. 

{\bf Test Conditions.}
For a fair comparison, we exactly follow the MPEG common test conditions (CTC)~\cite{MPEG_GPCC_CTC} to generate the anchor results using MPEG G-PCC.
The latest reference software TMC13-v14 of G-PCC~\cite{tmc13code} is used, and the G-PCC octree codec  is enabled in study. The augular coding mode is disabled when compressing the LiDAR point clouds based on the assumption that we do not utilize any prior knowledge from  point cloud acquisition stage.
For the lossless mode of G-PCC, the ``positionQuantizationScale'' is set to 1, while for the lossy compression mode, it is set to  1/2, 1/4, 1/8, etc., to offer variable bit rates. 
When evaluating the distortion for lossy compression, both {\it point-to-point error} (D1) and {\it point-to-plane error} (D2) are used to derive the PSNR. For lossless compression, the {\it bits per point} (bpp) is used to measure the compression ratio. 

Our model prototype is implemented  using  PyTorch  and MinkowskiEngine~\cite{MinkowskiEngine}, which is tested  on  a computer  with  an  Intel  Xeon Silver 4210 CPU and an  Nvidia  GeForce  RTX  2080  GPU. 
We record the encoding  and  decoding time  following the methodology used in  G-PCC.  Because of the implementation variation, e.g., CPU vs. GPU, C/C++ vs. Python, the runtime comparison only  serves as the intuitive reference to have a general idea about the computational complexity.

\subsection{SparsePCGC Presets}

The SparsePCGC offers a variety of options to compress the input PCG by applying different settings of SOPA and SLNE models. This section exemplifies the typical configuration for subsequent performance evaluation and comparison.

\input{table/table_kernels.tex}

\subsubsection{Lossless Mode}
For either dense or sparse point clouds, the SparsePCGC applies the lossless mode for all scales to mandate the perfect reconstruction.
For simplicity, we exemplify the trade-off between the computational complexity and compression efficiency by encoding/decoding dense object point clouds in Table~\ref{table:models}. Note that computational complexity is presented using the encoding/decoding time averaged per frame. 

{\bf Choice of SOPA Model.}
When the SLNE function is not enabled in the encoder, the coding gain over the G-PCC anchor increases greatly from the toy baseline using One-Stage SOPA with 5.5\% loss, to the case using 8-Stage SOPA model with 38.5\% compression gain. As also detailed in the supplemental material,  $n$-Stage SOPA provides slightly better gain, e.g., 39.3\% through the use of autoregressive neighbors in a sequential order. However, the decoding time of $n$-Stage SOPA is unbearable, e.g., $\approx$2 hrs, due to the causal dependency in computation from one voxel to another.

As seen, the proposed 3-Stage or 8-Stage SOPA model not only provides the encouraging compression efficiency, but also just requires 1$\sim$2 seconds for encoding or decoding a PCG frame averaged among all 8iVFB test frames in Table~\ref{table:models}. Quantitatively, our SparsePCGC is faster than the G-PCC anchor, e.g., 1.86s vs. 5.88s of encoding and 1.78s vs. 3.34s of decoding. But these numbers are just used as a reference to indicate that the proposed SparsePCGC is a solution with relatively low computational complexity requirement. This is because C++/C implemented G-PCC mainly runs on CPU while our method prototyped using PyTorch framework runs on  GPUs (e.g., RTX 2080 used in comparison).

\input{table/table_main}
\begin{figure*}[t]
	\centering
	\subfloat[]{
	\includegraphics[width=1.62in]{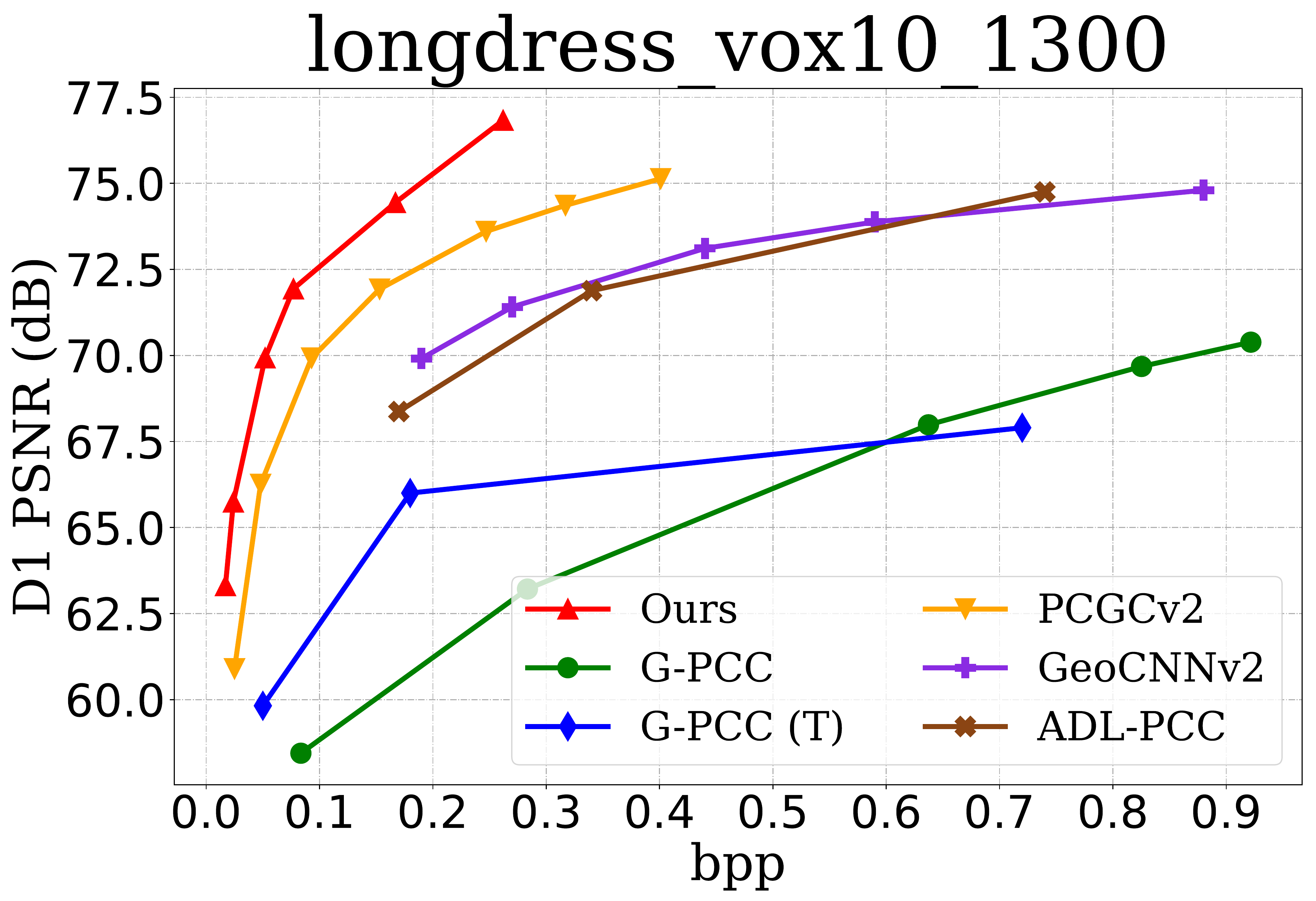}
	\includegraphics[width=1.62in]{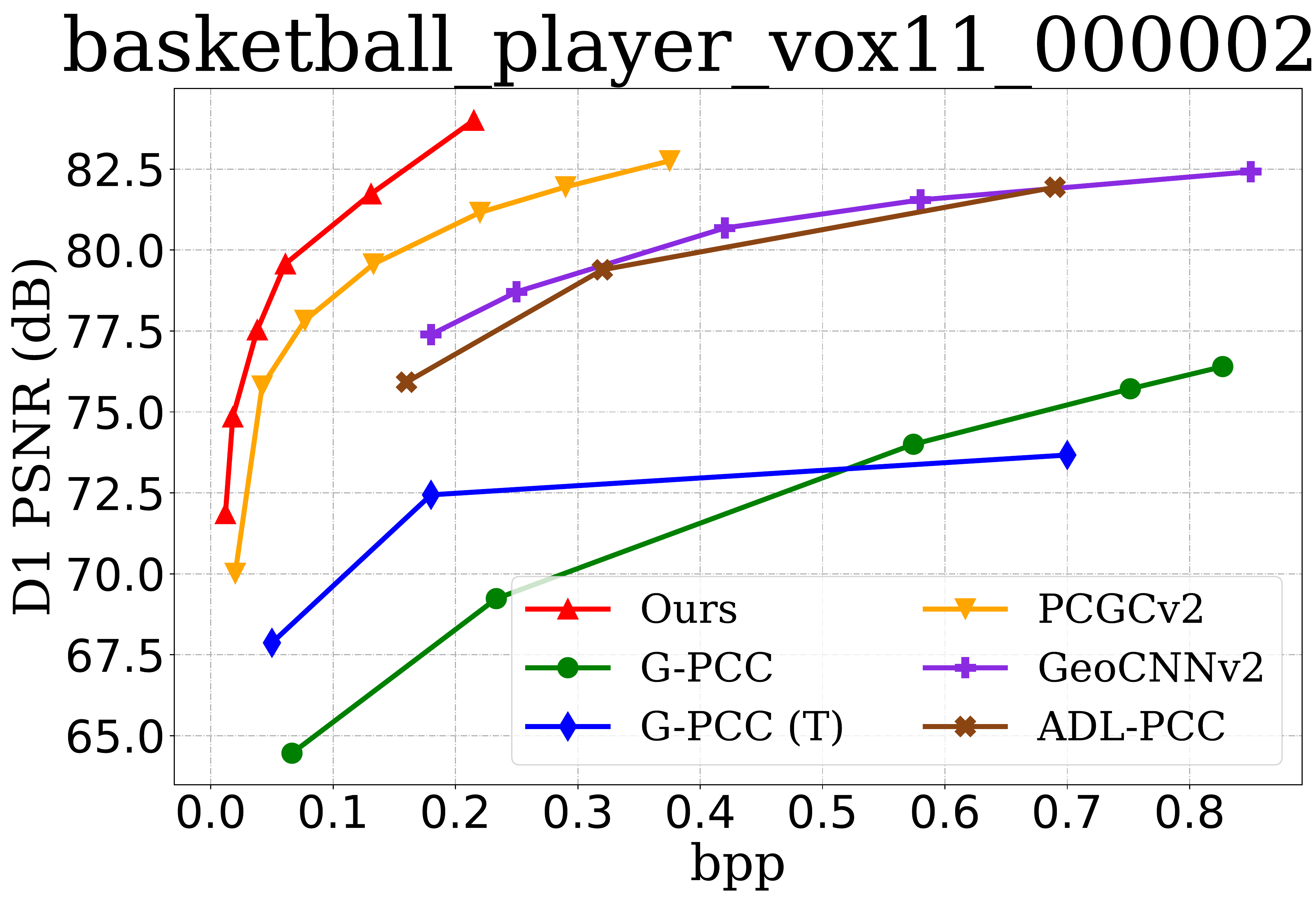}
	\includegraphics[width=1.62in]{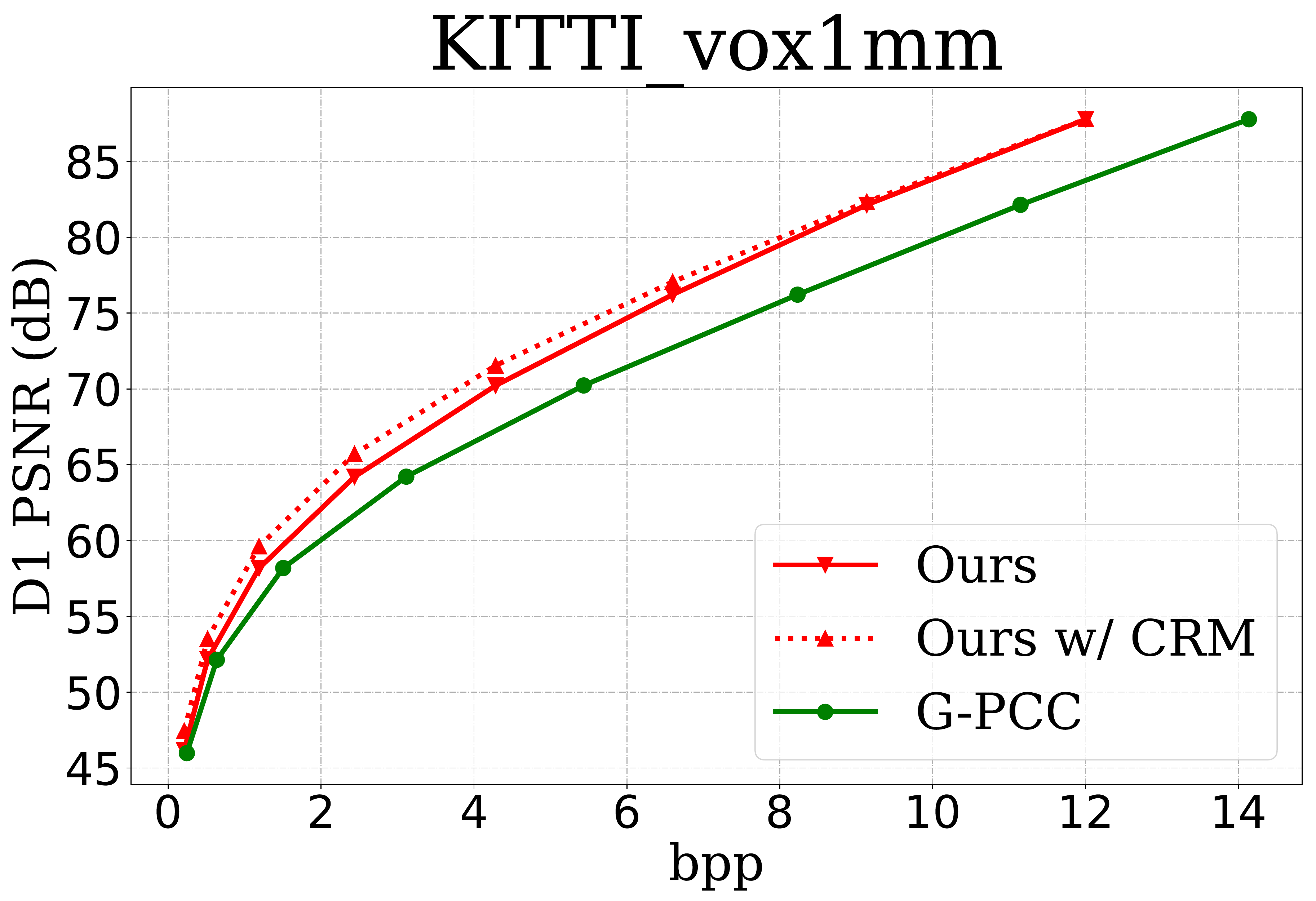}
	\includegraphics[width=1.62in]{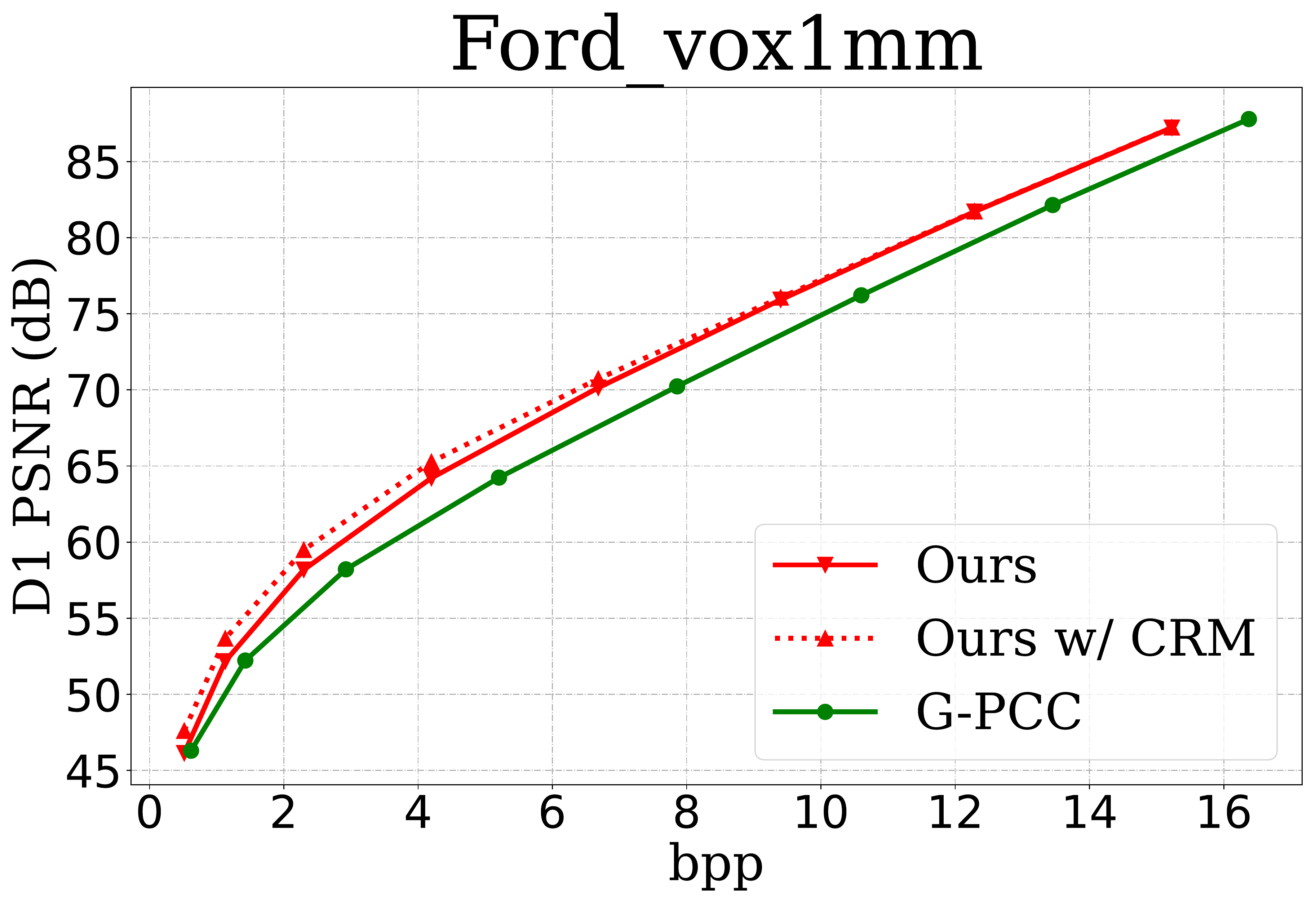}
	}
	\\
	\subfloat[]{
\includegraphics[width=1.62in]{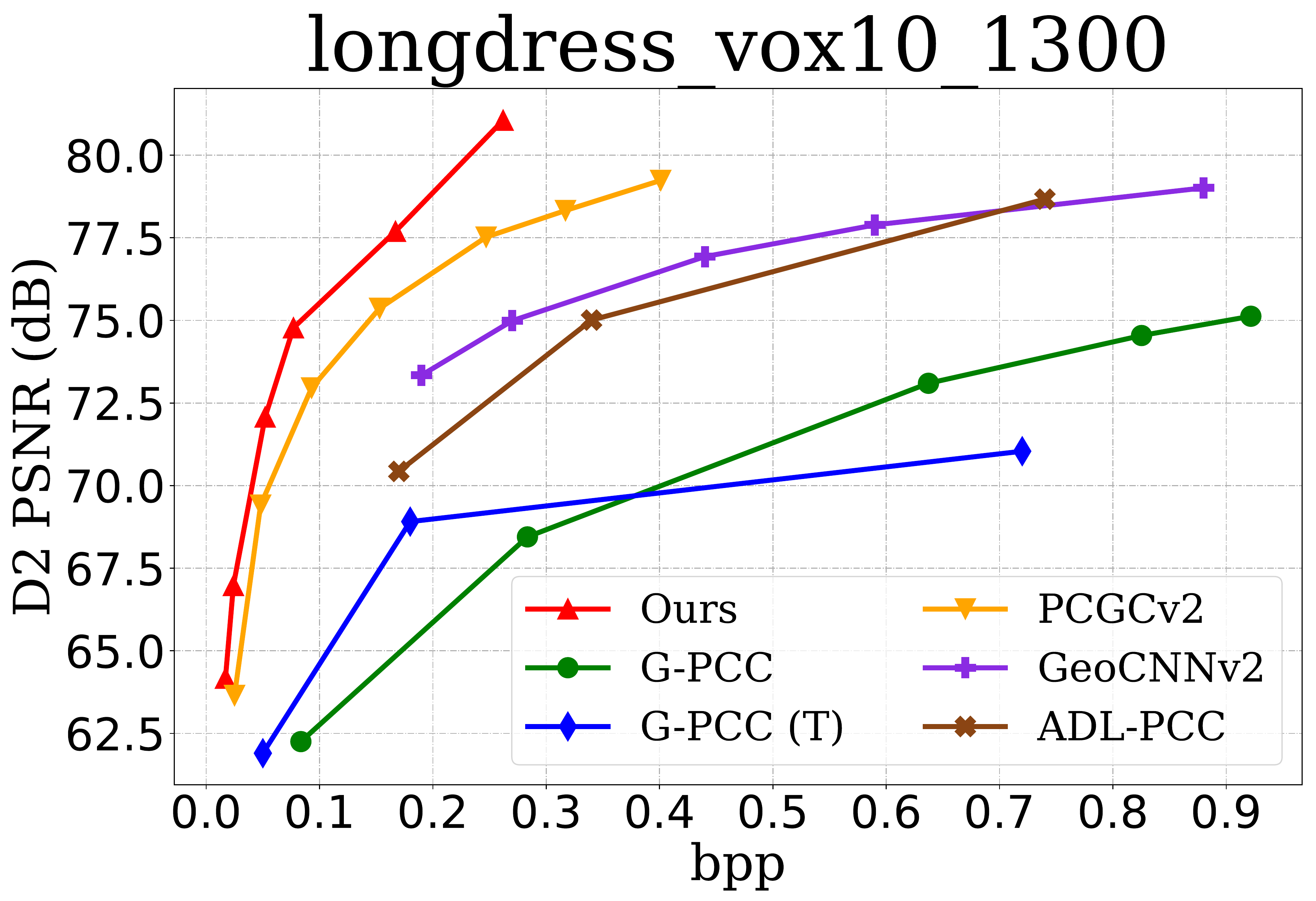}
	\includegraphics[width=1.62in]{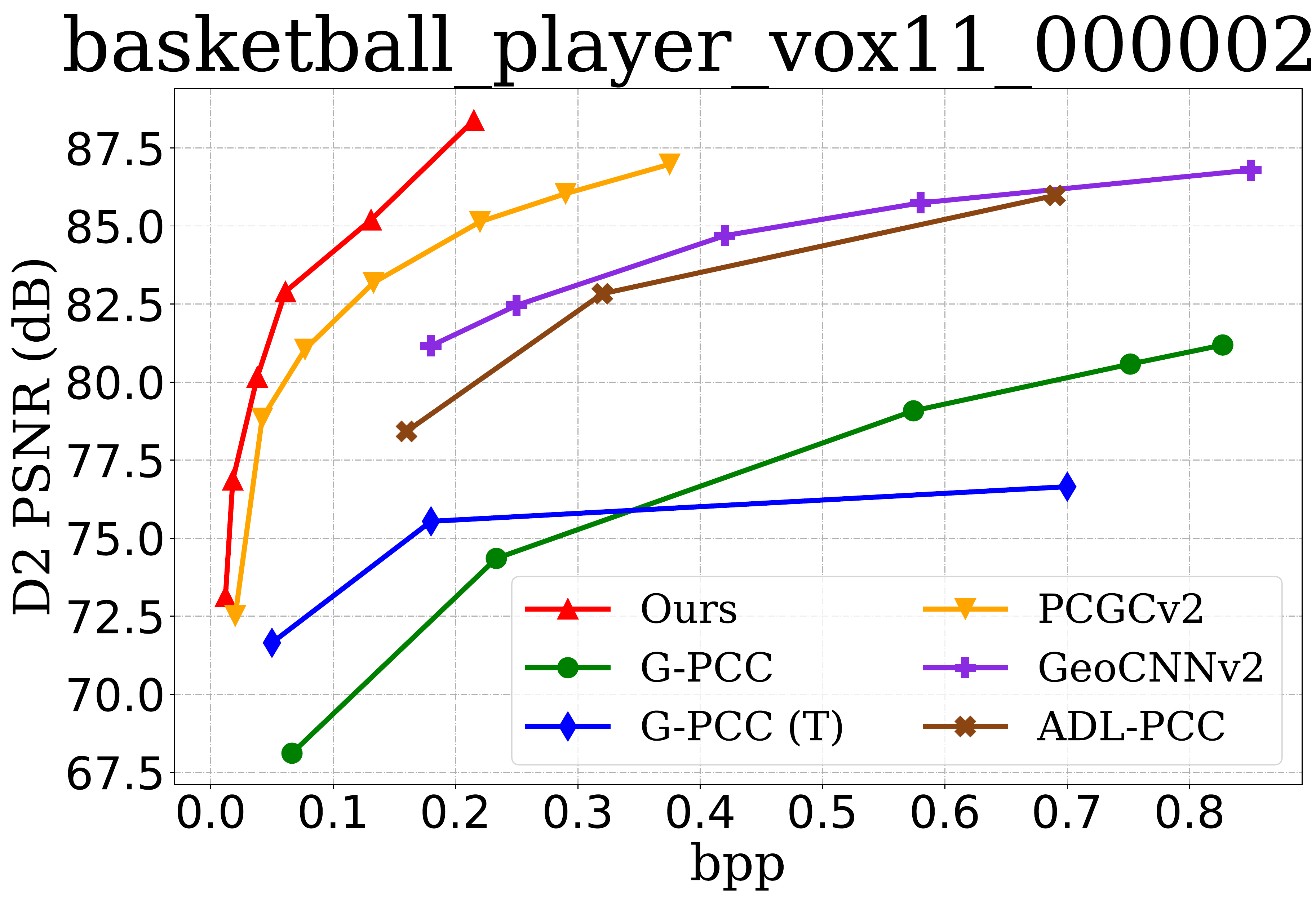}
	\includegraphics[width=1.62in]{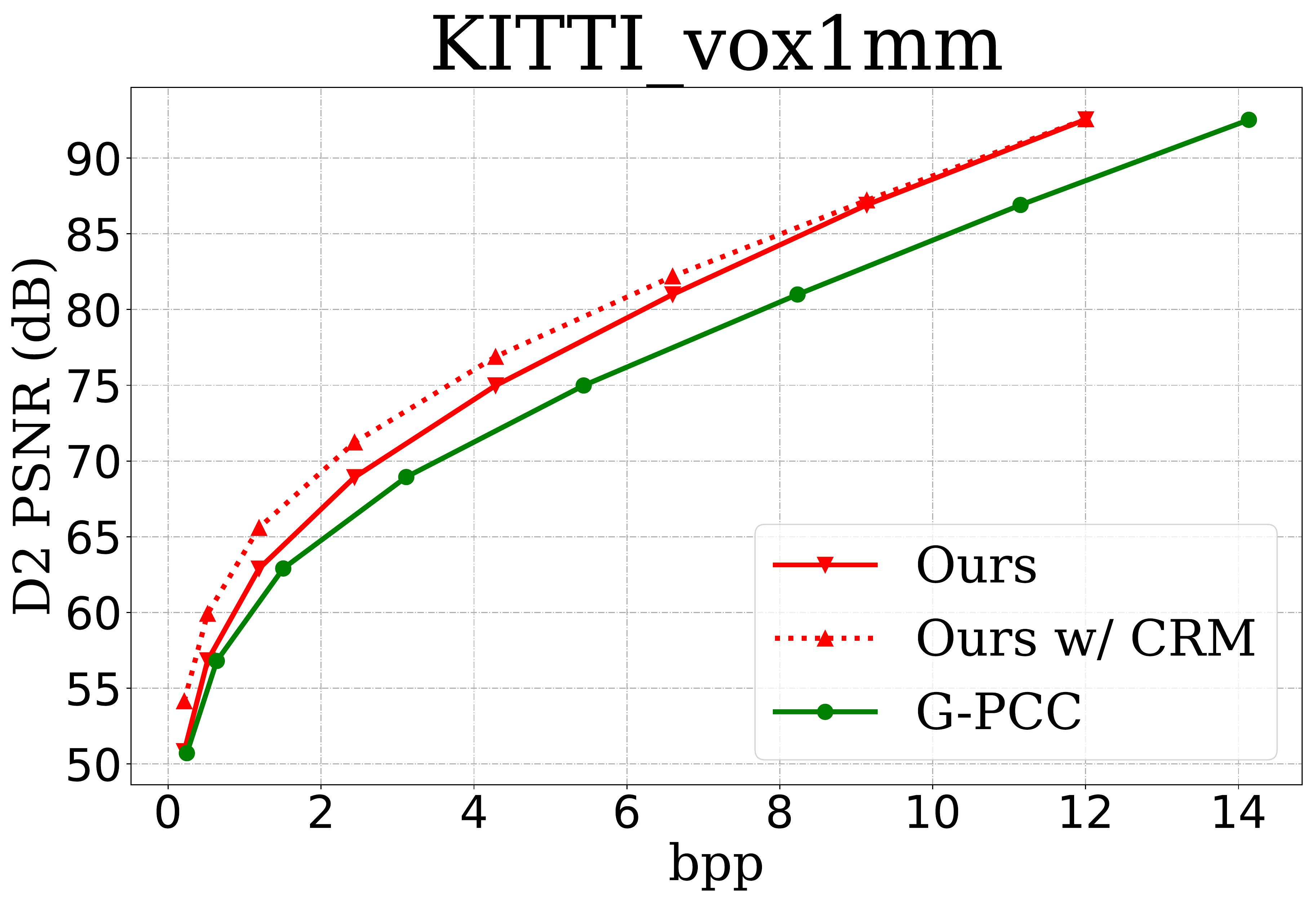}
	\includegraphics[width=1.62in]{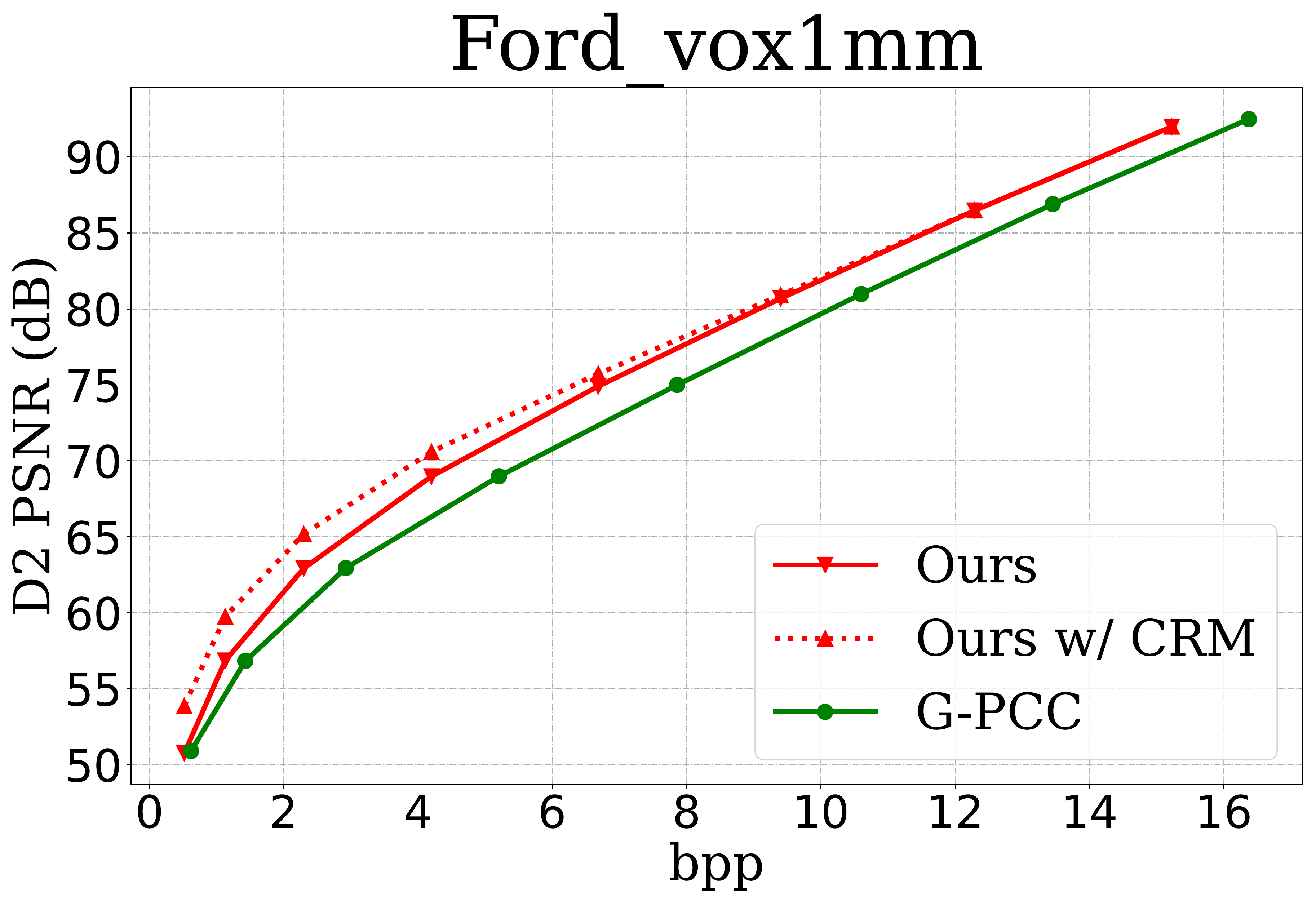}}
	\caption{{Performance comparison using rate-distortion curves. (a) D1 PSNR; (b) D2 PSNR.  G-PCC anchor using octree codec, G-PCC (T) using trisoup codec, {PCGCv2}~\cite{Wang2020MultiscalePC}, {GeoCNNv2}~\cite{Quach2020ImprovedDP}, and {ADL-PCC}~\cite{Guarda2021AdaptiveDL} are all compared.}}
	\label{fig:rdcurves}
\end{figure*}
We then explore the potential to use SLNE in the encoder to aggregate local spatial features for better probability estimation in SOPA model. Having the SLNE combined with  One-Stage SOPA, the coding efficiency is boosted from 5.5\% loss to 25.1\% gain, providing evidence to the effectiveness of the SLNE module. The coding gain can be further improved to 35.3\% when we combine the SLNE with 3-Stage SOPA, but the relative improvement from 3-Stage SOPA only is limited (e.g., from 31.7\% to 35.3\%). This relative improvement is negligible when comparing the SLNE with 8-Stage SOPA to the 8-Stage SOPA only. This occurs because the SLNE and multi-stage SOPA apply the similar mechanism to characterize and exploit local neighborhood correlations. Similarly, either ``SLNE enhanced One-Stage SOPA'' or ``SLNE enhanced 3-Stage SOPA'' shows low computational complexity by requiring less than 2 seconds to encode or decode a PCG frame on average.

As seen, 8-Stage SOPA and SLNE enhanced 3-Stage SOPA are promising candidates for lossless SparsePCGC because of their low complexity and superior compression performance. But training SOPA model only is much faster than training SLNE and SOPA jointly (about 2$\times$ speedup) from our extensive simulations.  To this end, ``8-Stage SOPA'' is used as the lossless mode of the proposed SparsePCGC.

{Previous discussions assume the settings of $k =3$ and $C = 32$ in SOPA {model} for the lossless compression of dense 8iVFB point clouds.} Next, we report the results when applying different $k$ and $C$ to compress  dense and sparse PCGs respectively. Note that we still enforce the lossless coding mode for simplicity.

{\bf Setting $k$ \& $C$.} We train four 8-Stage SOPA models by setting different $k$ and $C$ to evaluate the compression efficiency on both dense and sparse point clouds. In Table~\ref{table:kernel-size}, as for dense point clouds, the best result is reported when $k = 3$ and $C = 32$. Using the same $k = 3$, $C = 16$ is also a decent choice with less than 1 absolute percent point drop but results in greater than 70\% of model size reduction (4.89 MB vs. 1.31 MB). However, performance drops are observed when using a larger convolutional kernel, e.g., $k = 5$.  Such performance drop is not presented on the training dataset (ShapeNet), implying potential model overfitting with more parameters.

For sparse LiDAR point clouds, compression performance improves as $k$ and $C$ increase. About 10 absolute percentage points are reported when comparing the cases with $k=5$ and the cases with $k=3$ accordingly. We believe this is because the sparse nature of LiDAR points requires a larger convolutional kernel to capture sufficient neighbors  for effective information aggregation and embedding.

In a summary, the proposed SparsePCGC chooses to set $k$ = 3 and $C$ = 32 for dense point clouds, and $k$ = 5 and $C$ = 32 for sparse point clouds. 
In real-life applications, a preferred setting of $k$ and $C$ can be determined by balancing the complexity and performance tradeoff.

\subsubsection{Lossy Mode}\label{sec:mode_presets_lossy}

As in Fig.~\ref{fig:lossy_SparsePCGC}, in the lossless phase, we directly apply the 8-Stage SOPA model to exploit cross-scale and same-scale multi-stage correlations; while in the lossy phase, we use different methods for dense and sparse point clouds due to their diverse geometry precision and point density.
\begin{itemize}
    \item For dense PCG, we suggest the SLNE-enhanced One-Stage SOPA to upscale the sparse tensor from the $m$-th to the $(m+1)$-th scale and then apply the One-Stage SOPA from {$(m+1)$} till the highest scale $N$; 
    \item For sparse PCG, we apply the SOPA (Position) model to directly upscale the sparse tensor from the $m$-th scale to its highest scale $N$; 
\end{itemize}


\begin{figure}[t]
	\centering
	\subfloat[longdress]{\includegraphics[width=3.4in]{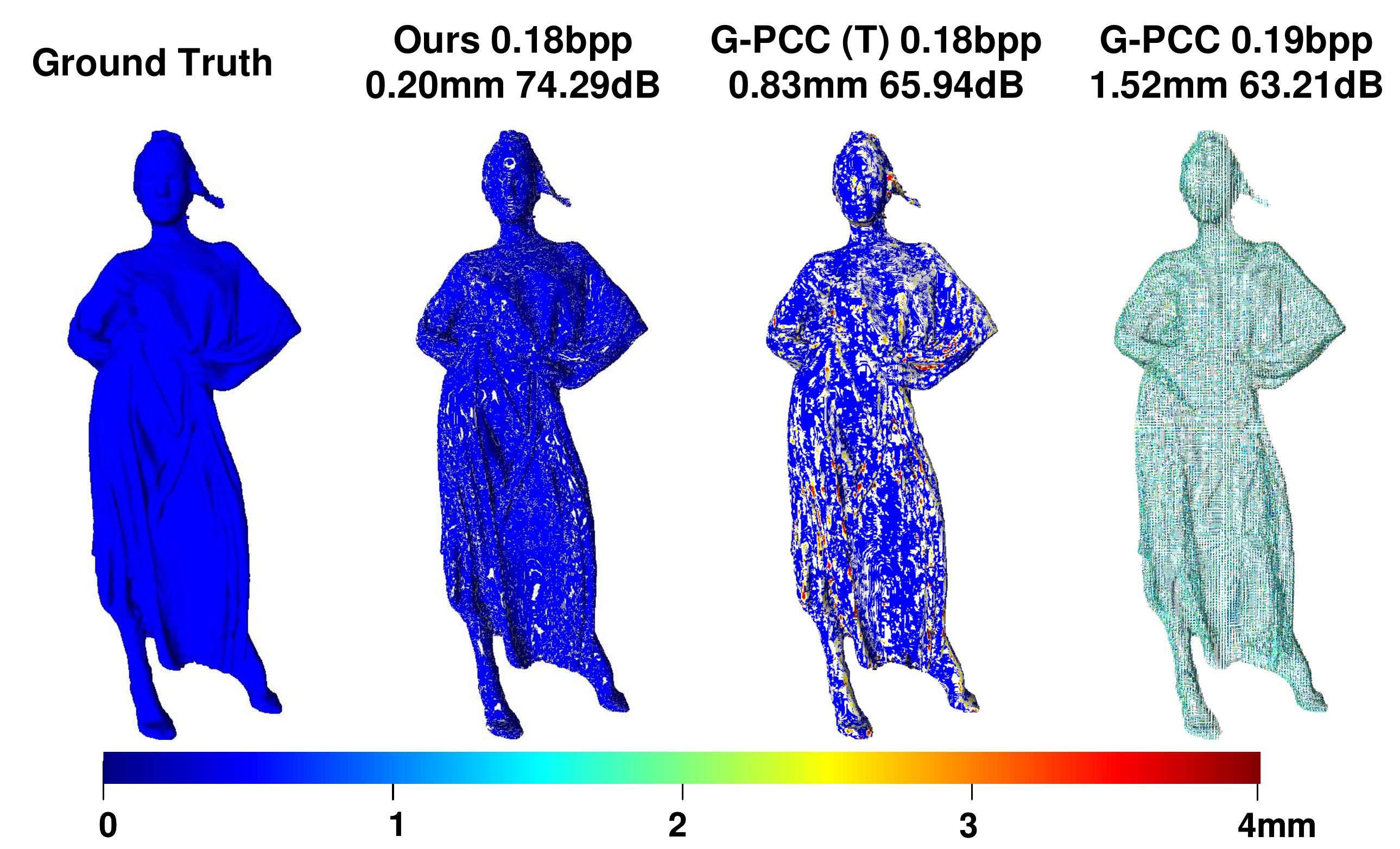}}\\
	\subfloat[Ford]{\includegraphics[width=3.4in]{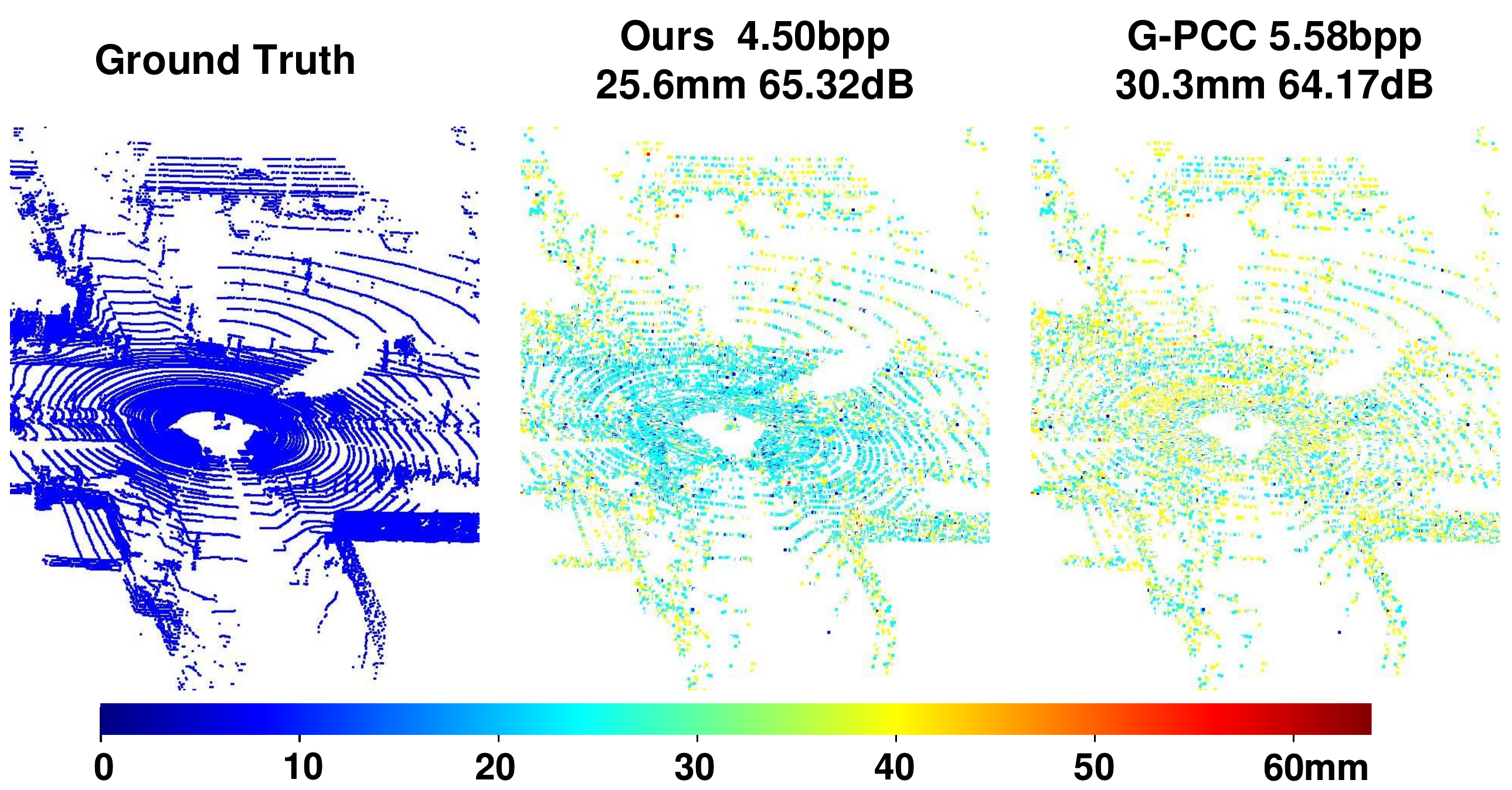}}
	\caption{Qualitative visualization of reconstruction of ``longdress\_vox10\_1300'' and ``Ford\_01\_vox1mm\_0500'' for ground truth, Ours, and G-PCC anchor. The color error map describes the point-to-point distortion measured in mm, and the numbers above represent the bitrate, mean error measured in mm, and D1 PSNR.}
	\label{fig:vis}
\end{figure}

Recalling that $m$ is adapted for various rate-distortion trade-off of lossy SparsePCGC, in our experiments, we set the factor $m$ as \{\textit{N}-1, \textit{N}-2, \textit{N}-3\} for dense point clouds, and \{\textit{N}-2, \textit{N}-3, $\cdots$, \textit{N}-9\} for sparse LiDAR point clouds.

%% file: table/table_multistage.tex
\begin{table*}[t]\small
\caption{{{Compression performance measured by bits per point (bpp) and complexity (runtime in seconds) tradeoff for various SOPA models  in lossless mode using 8iVFB samples. Encoding/decoding time is averaged for each frame.} $k$ = 3 \& $C$ = 32 in sparse convolution.}}
\label{table:models}
\centering
\begin{tabular}{|c|c|c|ccc|cc|}
\hline
\textbf{PCGs}        & \textbf{G-PCC} & \textbf{One-stage} & \textbf{3-Stage} & \textbf{8-Stage} & \textbf{n-Stage} & \textbf{\begin{tabular}[c]{@{}c@{}}SLNE Enh.\\ One-Stage\end{tabular}} & \textbf{\begin{tabular}[c]{@{}c@{}}SLNE Enh.\\ 3-Stage\end{tabular}} \\ \hline
Longdress            & 1.015          & 1.015              & 0.69             & 0.623            & 0.608            & 0.752                                                                  & 0.677                                                                \\
Loot                 & 0.970          & 1.032              & 0.66             & 0.594            & 0.581            & 0.713                                                                  & 0.546                                                                \\
Red\&balck           & 1.010          & 1.156              & 0.758            & 0.688            & 0.69             & 0.854                                                                  & 0.756                                                                \\
Soldier              & 1.030          & 1.103              & 0.702            & 0.627            & 0.62             & 0.764                                                                  & 0.684                                                                \\ \hline
\textbf{Average}     & 1.029          & 1.086              & 0.703            & 0.633            & \textbf{0.625}   & 0.77                                                                   & 0.666                                                                \\
\textbf{Gain}        & -              & +5.5\%             & -31.7\%          & -38.5\%          & \textbf{-39.3\%} & -25.1\%                                                                & -35.3\%                                                              \\ \hline
\textbf{EncTime (s)} & 5.88           & \textbf{0.66}      & 1.09             & 1.86             & 0.72             & 1.17                                                                   & 1.54                                                                 \\
\textbf{DecTime (s)} & 3.34           & \textbf{0.65}      & 1.00             & 1.78             & 2 hr             & 0.92                                                                   & 1.18                                                                 \\ \hline
\end{tabular}
\end{table*}

%% file: table/table_kernels.tex
\begin{table}[htb]\footnotesize
\caption{Impact of $k$ and $C$ used in sparse convolution for lossless compression examples.}
\label{table:kernel-size}
\centering
\begin{tabular}{|cccccc|}
\hline
\multicolumn{1}{|c|}{\textbf{\begin{tabular}[c]{@{}c@{}}kernel size\\ channels\end{tabular}}} & \multicolumn{1}{c|}{\textbf{G-PCC}} & \multicolumn{1}{c|}{\textbf{\begin{tabular}[c]{@{}c@{}}k3\\ C16\end{tabular}}} & \multicolumn{1}{c|}{\textbf{\begin{tabular}[c]{@{}c@{}}k3\\ C32\end{tabular}}} & \multicolumn{1}{c|}{\textbf{\begin{tabular}[c]{@{}c@{}}k5\\ C16\end{tabular}}} & \textbf{\begin{tabular}[c]{@{}c@{}}k5\\ C32\end{tabular}} \\ \hline
\multicolumn{1}{|c|}{\textbf{\begin{tabular}[c]{@{}c@{}}Model \\ Size (MB)\end{tabular}}}      & \multicolumn{1}{c|}{-}              & \multicolumn{1}{c|}{1.31}                                                      & \multicolumn{1}{c|}{4.89}                                                      & \multicolumn{1}{c|}{5.53}                                                      & 21.75                                                     \\ \hline
\multicolumn{6}{|c|}{\textbf{Dense PCGs (8iVFB\_vox10)}}                                                                                                                                                                                                                                                                                                                                                                                            \\ \hline
\multicolumn{1}{|c|}{\textbf{bpp}}                                                            & \multicolumn{1}{c|}{1.029}          & \multicolumn{1}{c|}{0.642}                                                     & \multicolumn{1}{c|}{\textbf{0.633}}                                            & \multicolumn{1}{c|}{0.642}                                                     & 0.655                                                     \\
\multicolumn{1}{|c|}{\textbf{gain}}                                                           & \multicolumn{1}{c|}{-}              & \multicolumn{1}{c|}{-37.6\%}                                                   & \multicolumn{1}{c|}{\textbf{-38.5\%}}                                          & \multicolumn{1}{c|}{-37.6\%}                                                   & -36.3\%                                                   \\ \hline
\multicolumn{1}{|c|}{\textbf{EncTime (s)}}                                                     & \multicolumn{1}{c|}{5.88}           & \multicolumn{1}{c|}{\textbf{1.41}}                                             & \multicolumn{1}{c|}{1.86}                                                      & \multicolumn{1}{c|}{2.45}                                                      & 3.44                                                      \\
\multicolumn{1}{|c|}{\textbf{DecTime (s)}}                                                     & \multicolumn{1}{c|}{3.34}           & \multicolumn{1}{c|}{\textbf{1.28}}                                             & \multicolumn{1}{c|}{1.78}                                                      & \multicolumn{1}{c|}{2.45}                                                      & 3.31                                                      \\ \hline
\multicolumn{6}{|c|}{\textbf{Sparse PCGs (KITTI\_vox2cm)}}                                                                                                                                                                                                                                                                                                                                                                                          \\ \hline
\multicolumn{1}{|c|}{\textbf{bpp}}                                                            & \multicolumn{1}{c|}{7.637}          & \multicolumn{1}{c|}{7.021}                                                     & \multicolumn{1}{c|}{6.880}                                                      & \multicolumn{1}{c|}{6.334}                                                      & \textbf{6.130}                                             \\
\multicolumn{1}{|c|}{\textbf{gain}}                                                           & \multicolumn{1}{c|}{\textbf{-}}     & \multicolumn{1}{c|}{-8.1\%}                                                    & \multicolumn{1}{c|}{-9.9\%}                                                    & \multicolumn{1}{c|}{-17.1\%}                                                   & \textbf{-19.7\%}                                          \\ \hline
\multicolumn{1}{|c|}{\textbf{EncTime (s)}}                                                     & \multicolumn{1}{c|}{1.16}           & \multicolumn{1}{c|}{\textbf{1.15}}                                             & \multicolumn{1}{c|}{1.23}                                                      & \multicolumn{1}{c|}{1.40}                                                       & 1.73                                                      \\
\multicolumn{1}{|c|}{\textbf{DecTime (s)}}                                                     & \multicolumn{1}{c|}{0.72}           & \multicolumn{1}{c|}{1.04}                                                      & \multicolumn{1}{c|}{1.12}                                                      & \multicolumn{1}{c|}{1.32}                                                      & 1.61                                                      \\ \hline
\end{tabular}
\end{table}

%% file: table/table_main.tex
\begin{table*}[htb]\small
\caption{{Quantitative performance gains to the G-PCC {anchor}. BD-Rate measurement is used for lossy mode and compression ratio (CR) gain measured by bits per point (bpp) is used  in lossless mode.}}
\label{table:gpcc}
\centering
\begin{tabular}{|cccccccccc|}
\hline
\multicolumn{1}{|c|}{\multirow{3}{*}{\textbf{Point Clouds}}} & \multicolumn{5}{c|}{\textbf{lossless}}                                                                                                                                                                                                                                                                               & \multicolumn{4}{c|}{\textbf{lossy}}                                                                                                                                                                                       \\ \cline{2-10} 
\multicolumn{1}{|c|}{}                                       & \multicolumn{2}{c|}{\textbf{G-PCC}}                                                                                          & \multicolumn{2}{c|}{\textbf{Ours}}                                                                                              & \multicolumn{1}{c|}{\multirow{3}{*}{\textbf{CR Gain}}} & \multicolumn{1}{c|}{\textbf{G-PCC}}                                                      & \multicolumn{1}{c|}{\textbf{Ours}}                                                       & \multicolumn{2}{c|}{\textbf{BD-Rate Gain}} \\ \cline{2-5} \cline{7-10}
\multicolumn{1}{|c|}{}                                       & \multicolumn{1}{c|}{\textbf{bpp}} & \multicolumn{1}{c|}{\textbf{\begin{tabular}[c]{@{}c@{}}Time (s)\\ Enc/Dec\end{tabular}}} & \multicolumn{1}{c|}{\textbf{bpp}}    & \multicolumn{1}{c|}{\textbf{\begin{tabular}[c]{@{}c@{}}Time (s)\\ Enc/Dec\end{tabular}}} & \multicolumn{1}{c|}{}                               & \multicolumn{1}{c|}{\textbf{\begin{tabular}[c]{@{}c@{}}Time (s)\\ Enc/Dec\end{tabular}}} & \multicolumn{1}{c|}{\textbf{\begin{tabular}[c]{@{}c@{}}Time (s)\\ Enc/Dec\end{tabular}}} & \textbf{D1}      & \textbf{D2}      \\ \hline
\multicolumn{10}{|c|}{\textbf{Dense Object}}                                                                                                                                                                                                                                                                                                                                                                                                                                                                                                                                                                           \\ \hline
\multicolumn{1}{|c|}{longdress\_vox10}                       & \multicolumn{1}{c|}{1.015}        & \multicolumn{1}{c|}{5.73/3.22}                                                           & \multicolumn{1}{c|}{0.625}           & \multicolumn{1}{c|}{1.80/1.70}                                                           & \multicolumn{1}{c|}{-38.4\%}                        & \multicolumn{1}{c|}{4.84/2.50}                                                           & \multicolumn{1}{c|}{0.68/1.19}                                                           & -94.5\%          & -89.7\%          \\
\multicolumn{1}{|c|}{loot\_vox10}                            & \multicolumn{1}{c|}{0.970}        & \multicolumn{1}{c|}{5.36/3.01}                                                           & \multicolumn{1}{c|}{0.596}           & \multicolumn{1}{c|}{1.75/1.60}                                                           & \multicolumn{1}{c|}{-38.6\%}                        & \multicolumn{1}{c|}{4.56/2.64}                                                           & \multicolumn{1}{c|}{0.64/1.10}                                                           & -95.0\%          & -90.3\%          \\
\multicolumn{1}{|c|}{red\&black\_vox10}                      & \multicolumn{1}{c|}{1.100}        & \multicolumn{1}{c|}{6.72/3.56}                                                           & \multicolumn{1}{c|}{0.690}           & \multicolumn{1}{c|}{1.68/1.84}                                                           & \multicolumn{1}{c|}{-37.3\%}                        & \multicolumn{1}{c|}{4.07/1.89}                                                           & \multicolumn{1}{c|}{0.64/1.08}                                                           & -93.6\%          & -88\%            \\
\multicolumn{1}{|c|}{soldier\_vox10}                         & \multicolumn{1}{c|}{1.030}        & \multicolumn{1}{c|}{5.07/3.04}                                                           & \multicolumn{1}{c|}{0.628}           & \multicolumn{1}{c|}{2.22/1.99}                                                           & \multicolumn{1}{c|}{-39.0\%}                        & \multicolumn{1}{c|}{5.65/3.32}                                                           & \multicolumn{1}{c|}{0.78/1.40}                                                           & -94.3\%          & -89.2\%          \\
\multicolumn{1}{|c|}{queen\_vox10}                           & \multicolumn{1}{c|}{0.773}        & \multicolumn{1}{c|}{7.34/4.09}                                                           & \multicolumn{1}{c|}{0.547}           & \multicolumn{1}{c|}{1.92/1.85}                                                           & \multicolumn{1}{c|}{-29.2\%}                        & \multicolumn{1}{c|}{4.97/2.51}                                                           & \multicolumn{1}{c|}{0.66/1.19}                                                           & -95\%            & -90.1\%          \\
\multicolumn{1}{|c|}{player\_vox11}                          & \multicolumn{1}{c|}{0.898}        & \multicolumn{1}{c|}{23.66/14.95}                                                         & \multicolumn{1}{c|}{0.520}           & \multicolumn{1}{c|}{6.43/6.07}                                                           & \multicolumn{1}{c|}{-42.1\%}                        & \multicolumn{1}{c|}{16.05/5.99}                                                          & \multicolumn{1}{c|}{1.12/2.49}                                                           & -97.2\%          & -94.9\%          \\
\multicolumn{1}{|c|}{dancer\_vox11}                          & \multicolumn{1}{c|}{0.880}        & \multicolumn{1}{c|}{19.45/12.63}                                                         & \multicolumn{1}{c|}{0.514}           & \multicolumn{1}{c|}{5.99/5.67}                                                           & \multicolumn{1}{c|}{-41.6\%}                        & \multicolumn{1}{c|}{14.74/6.82}                                                          & \multicolumn{1}{c|}{1.17/2.32}                                                           & -96.9\%          & -93.9\%          \\ \hline
\multicolumn{1}{|c|}{\textbf{Average}}                       & \multicolumn{1}{c|}{0.952}        & \multicolumn{1}{c|}{10.48/6.36}                                                          & \multicolumn{1}{c|}{\textbf{0.589}}  & \multicolumn{1}{c|}{\textbf{3.11/2.96}}                                                  & \multicolumn{1}{c|}{\textbf{-38.2\%}}               & \multicolumn{1}{c|}{7.84/3.81}                                                           & \multicolumn{1}{c|}{\textbf{0.81/1.54}}                                                  & \textbf{-95.2\%} & \textbf{-90.9\%} \\ \hline
\multicolumn{10}{|c|}{\textbf{Sparse LiDAR}}                                                                                                                                                                                                                                                                                                                                                                                                                                                                                                                                                                          \\ \hline
\multicolumn{1}{|c|}{KITTI\_q2cm}                            & \multicolumn{1}{c|}{7.637}        & \multicolumn{1}{c|}{1.16/0.72}                                                           & \multicolumn{1}{c|}{6.130}           & \multicolumn{1}{c|}{1.73/1.61}                                                           & \multicolumn{1}{c|}{-19.7\%}                        & \multicolumn{1}{c|}{0.85/0.49}                                                           & \multicolumn{1}{c|}{1.26/0.89}                                                           & -33.4\%          & -44.2\%          \\
\multicolumn{1}{|c|}{Ford\_q2mm}                             & \multicolumn{1}{c|}{10.032}       & \multicolumn{1}{c|}{0.96/0.59}                                                           & \multicolumn{1}{c|}{8.784}           & \multicolumn{1}{c|}{1.59/1.47}                                                           & \multicolumn{1}{c|}{-12.4\%}                        & \multicolumn{1}{c|}{0.75/0.55}                                                           & \multicolumn{1}{c|}{1.31/0.99}                                                           & -29.8\%          & -37.6\%          \\ \hline
\multicolumn{1}{|c|}{KITTI\_q1mm}                            & \multicolumn{1}{c|}{20.154}       & \multicolumn{1}{c|}{1.77/1.03}                                                           & \multicolumn{1}{c|}{17.971}          & \multicolumn{1}{c|}{2.91/1.71}                                                           & \multicolumn{1}{c|}{-10.8\%}                        & \multicolumn{1}{c|}{1.26/0.72}                                                           & \multicolumn{1}{c|}{1.70/1.28}                                                           & -29.1\%          & -36.0\%          \\
\multicolumn{1}{|c|}{Ford\_q1mm}                             & \multicolumn{1}{c|}{22.298}       & \multicolumn{1}{c|}{1.24/0.74}                                                           & \multicolumn{1}{c|}{21.168}          & \multicolumn{1}{c|}{2.65/2.47}                                                           & \multicolumn{1}{c|}{-5.1\%}                         & \multicolumn{1}{c|}{0.98/0.78}                                                           & \multicolumn{1}{c|}{1.72/1.31}                                                           & -22.0\%          & -26.8\%          \\ \hline
\multicolumn{1}{|c|}{\textbf{Average}}                       & \multicolumn{1}{c|}{15.030}       & \multicolumn{1}{c|}{\textbf{1.28/0.77}}                                                  & \multicolumn{1}{c|}{\textbf{13.513}} & \multicolumn{1}{c|}{\textbf{2.22/2.06}}                                                  & \multicolumn{1}{c|}{\textbf{-12.0\%}}               & \multicolumn{1}{c|}{\textbf{0.96/0.64}}                                                  & \multicolumn{1}{c|}{1.50/1.12}                                                           & \textbf{-28.6\%} & \textbf{-36.2\%} \\ \hline
\end{tabular}
\end{table*}

%% file: 4_exp_part2.tex
\subsection{Performance Evaluation and Comparison}

\subsubsection{Comparison to the G-PCC}

We first report performance gains of the proposed SparsePCGC to standardized G-PCC anchor in Table~\ref{table:gpcc}.

{\bf Compression Ratio (CR) Gain in Lossless Mode.}
For dense point clouds, our SparsePCGC improves the G-PCC by  {38.2}\% on average and up to  {42.1}\%  for basketball\_player\_vox11\_0200. For sparse point clouds, our method still outperforms the G-PCC,  although the gains are not as high as that of dense point clouds. This is because sparse point clouds exhibit much sparser spatial distribution and it is relatively difficult to capture and characterize inter-point correlations.

{\bf BD-Rate Improvement in Lossy Mode.}
We also evaluate the lossy compression  efficiency of our SparsePCGC to the anchor G-PCC.
On average, the proposed SparsePCGC shows {95.2}\% and {90.9}\% BD-Rate improvement for dense PCGs and {28.6\% and 36.2\%} gains for sparse PCGs, when the distortion is respectively measured by D1 and D2 metrics. 
Quantitative gains to the G-PCC anchor are also visualized in Fig.~\ref{fig:rdcurves} using rate-distortion curves.
Qualitative comparisons with G-PCC are presented in Fig.~\ref{fig:vis}, using point clouds colored by reconstruction error. All of these clearly demonstrate the superior efficiency of the proposed SparsePCGC.

\textbf{Runtime Evaluation}.
We also measure the encoding and decoding time for the G-PCC anchor and our method  in Table~\ref{table:gpcc}. 
Quantitatively, our encoding/decoding is about 2$\sim$3$\times$ faster than G-PCC on dense object point clouds, while about 2$\sim$3$\times$ slower than G-PCC on sparse LiDAR point clouds in lossless mode. A similar runtime trend is observed in lossy mode. The G-PCC is faster on sparse LiDAR PCGs probably because of its dedicated tools engineered to accelerate the process of isolated points~\cite{MPEG_PCC_PIEEE}, while our method currently treats all points equally in the same framework.  We particularly notice that our method reports more than 9$\times$ speedup to that of the G-PCC encoder when encoding dense PCGs in the lossy mode. This is because simple dyadic downscaling and fast One-Stage SOPA are used in the proposed SparsePCGC (see Fig.~\ref{fig:lossy_SparsePCGC}b).  
Overall, our SparsePCGC is a low-complexity solution, requiring only a few seconds to encode/decode large-scale point clouds with state-of-the-art performance. 

For G-PCC, we collect the runtime from the log file recorded in the reference software. For our method, we include the time for both network inference and entropy engine for bitstream handling. The G-PCC is implemented using C++/C and tested on CPU while our method is written in Python and tested on RTX 2080 GPU. It requires substantial efforts to take implementation details such as CPU vs. GPU, C++/C vs. Python, and code optimization levels into the consideration, which is deferred as our future study. Thus, in this work, the runtime measurement and comparison is just an intuitive reference to present a general idea of the computational complexity.

\input{table/table_voxeldnn}

\input{table/table_lossy}

\subsubsection{Comparison to other Learning-based Solutions} 
Most learning-based approaches can only deal with a single type of point clouds. Thus, we detail the comparison individually. Again, we  reiterate that the encoding/decoding inference  runs on  RTX 2080 GPU to best match the computation device used by other learning-based solutions. In terms of the model training, we also try our best to apply the common sets used by others for fair comparisons. More details are offered in our supplemental material.

{\bf Lossless Compression of Dense PCGs.}
Nguyen~\textit{et al.}~\cite{Nguyen2021LosslessCO} developed the VoxelDNN - a learning-based lossless compression method for dense PCG, in which masked 3D CNN is applied for voxel occupancy probability approximation assuming the use of a uniform voxel representation model. The VoxelDNN provided promising compression ratios at the expense of a very long encoding and decoding time due to the sequential processing. A multiscale parallel version of VoxelDNN, referred to as the MsVoxelDNN~\cite{Nguyen2021MultiscaleDC}, was then proposed to speed up the runtime (e.g., 12$\times$ for decoding, and 16$\times$ for encoding). We directly quote numbers from their papers. It is noted that only one frame is tested for VoxelDNN-related methods.

In contrast to the uniform voxel representation, Kaya~\textit{et al.}~\cite{Kaya2021NeuralNM} proposed an octree-based lossless compression method --- NNOC, in which it estimated the occupancy probability of the octree node based on available nodes surrounding it. Similarly, the NNOC also suffered from an extremely long decoding time. Then, a fast version of NNOC, a.k.a., fNNOC, was updated for decoding acceleration (e.g., about 10$\times$). We best reproduce the results using sources from NNOC/fNNOC for evaluation.

Another learning-based octree codec, a.k.a, OctAttention~\cite{Fu2022OctAttentionOL} that reported state-of-the-art efficiency through the use of Transformer to effectively characterize and embed correlations across a large amount of previously-processed neighbors (e.g., parent nodes, sibling nodes) for context modeling was included for comparison as well. The OctAttention achieved fast encoding with just 1$\sim$2 seconds because full data availability in the encoder allowed the use of massive encoding parallelism; however, its decoding was super slow due to causal data dependency. We reproduce the results of the OctAttention model using its publicly-accessible sources.

Table~\ref{table:learnedPCClossless} quantitatively compares the lossless compression efficiency of the proposed SparsePCGC, VoxelDNN (MsVoxelDNN), NNOC (fNNOC), and OctAttention. We test sequences in both 8iVFB and MVUB. Except for VoxelDNN solutions that encode/decode one frame for each test sequence (a.k.a, ``one''), all other solutions process all frames (a.k.a., ``seqs'') in each sequence for averaged  results\footnote{Only one frame is contained in both ``Thaidancer$\times$1'' and ``Boxer$\times$1'' sequences, implying the same results for  ``one'' and ``seqs'' categories.}.

On top of the same G-PCC anchor,  our method attains the best gains, e.g., on average, 44.2\% for 8iVFB  and 37.7\% for MVUB sequences. More than 20 absolute percentage points are captured over the MsVoxelDNN and fNNOC. As compared with the OctAttention that requires 1024 neighboring nodes for context modeling, our SparsePCGC that solely relies on local neighbors  within a limited receptive field (e.g., $3\times3\times3$ for dense PCG) provides another 7$\sim$10 absolute percentage points improvement approximately.

As for the encoding/decoding runtime comparison with other learning-based solutions, our method speeds up the MsVoxelDNN/fNNOC at least 20$\times$. Though the OctAttention model provides the fastest encoding, it is still at the same order of magnitude as our method, not to mention that its decoding time is more than 500$\times$ slower.

\input{table/table_lidar}

{\bf Lossy Compression of Dense PCGs.}
We then compare the SparsePCGC with the PCGCv2~\cite{Wang2020MultiscalePC}, GeoCNNv2~\cite{Quach2020ImprovedDP}, and ADL-PCC~\cite{Guarda2021AdaptiveDL} for lossy compression of dense PCGs.
Both GeoCNNv2~\cite{Quach2020ImprovedDP} and ADL-PCC~\cite{Guarda2021AdaptiveDL} leverage the dense CNNs assuming the use of uniform voxel representation.
The PCGCv2~\cite{Wang2020MultiscalePC} is our earlier attempt to apply the multiscale sparse tensor, which applies the fixed factors for resolution scaling, uses the lossless G-PCC to code geometry coordinates at the lowest scale, and supports the lossy compression of dense PCGs only.
Both rate-distortion curves in Fig.~\ref{fig:rdcurves} and BD-Rate gains in Table~\ref{table:BDBR_lossy_dense} evidence that the SparsePCGC presents significant performance lead over these methods. Note that lossy PCGC approaches are studied extensively in MPEG for the development of next-generation learning-based PCGC standard. A companion document~\cite{MPEG-AI_PCC} also reports the superior efficiency of the proposed method from the third-party evaluations.

\begin{figure}[b]
\centering
\subfloat[]{\includegraphics[width=1.65in]{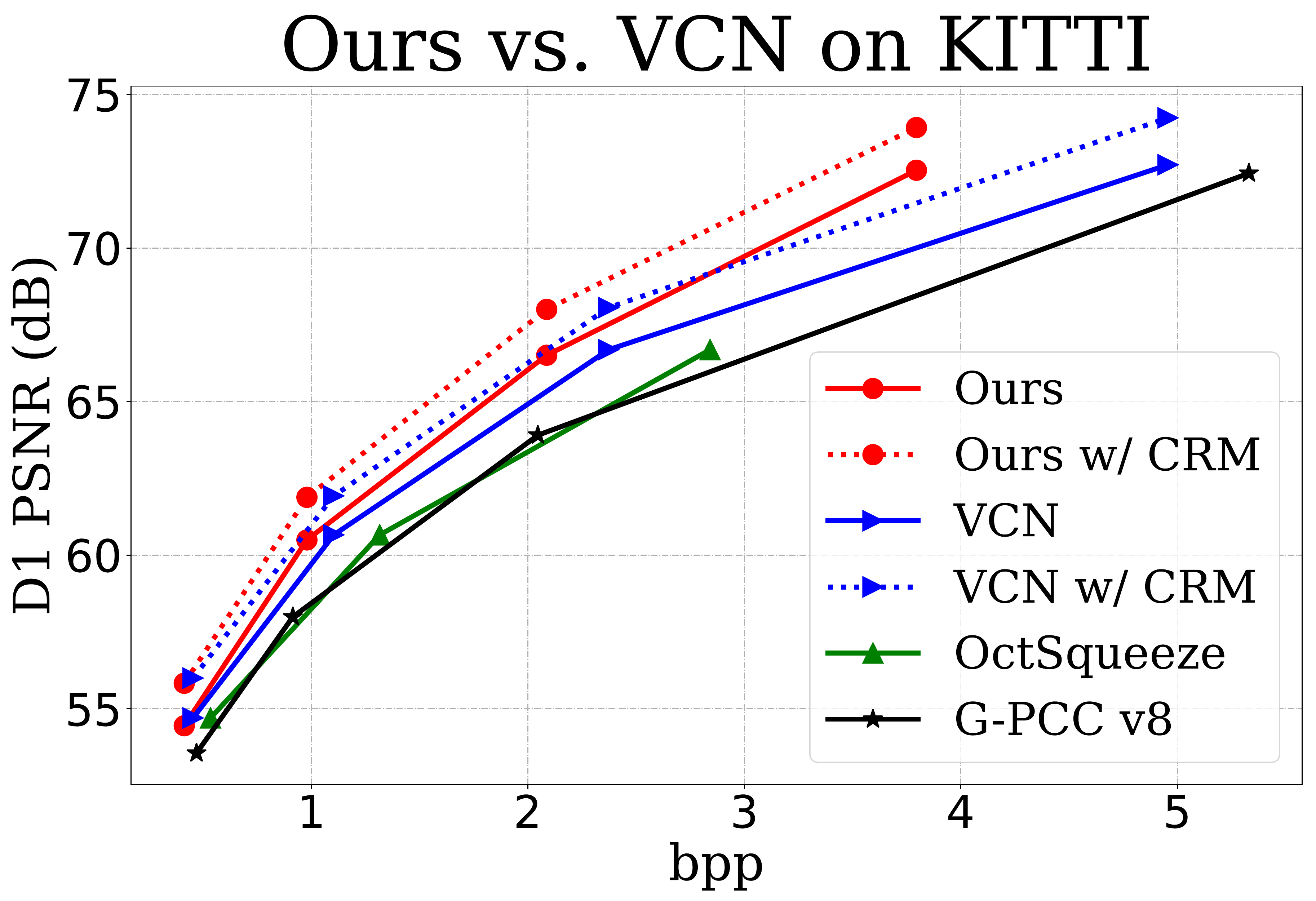} \label{fig:rd-vcn}}
\subfloat[]{\includegraphics[width=1.65in]{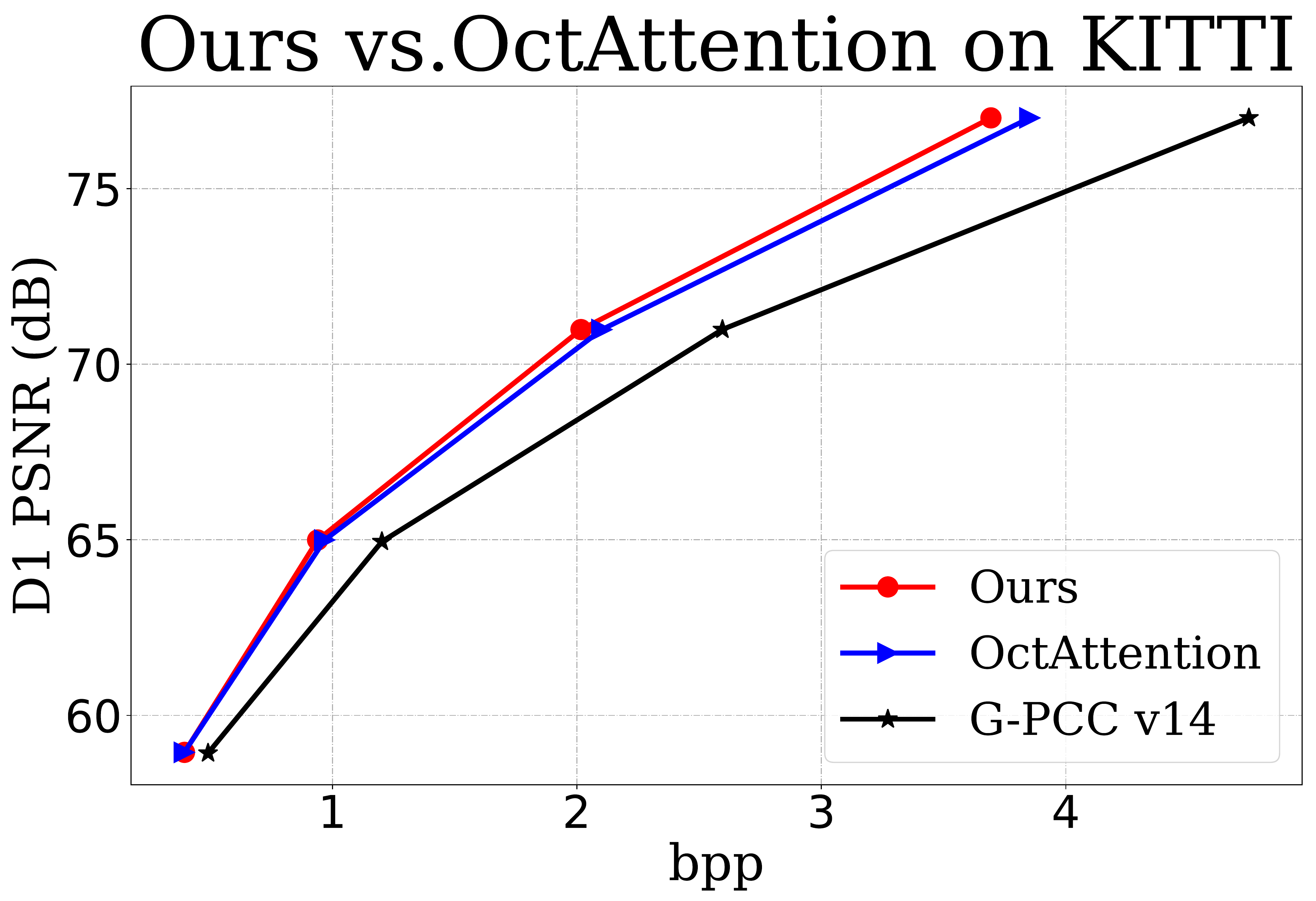}\label{fig:rd-attn}}
\caption{{Performance comparison using rate-distortion curves (a) Ours versus the VCN under the same test conditions used by VCN~\cite{Que2021VoxelContextNetAO}; (b) Ours versus the OctAttention under the same test conditions used in OctAttention~\cite{Fu2022OctAttentionOL}.}}
\label{fig:perf_VCN_OctAttention}
\end{figure}

\textbf{Lossy Compression of Sparse LiDAR PCGs.}
Most approaches adopt the octree model to represent sparse LiDAR point clouds. 
Huang {\it et al.}~\cite{Huang2020OctSqueezeOE} proposed the OctSqueeze that used MLP to exploit the dependency between parent and child nodes. Que {\it et al.}~\cite{Que2021VoxelContextNetAO} then proposed the VCN (VoxelContext-Net) which arranged sibling nodes using the uniform voxel model and employed 3D CNN to exploit the dependency across nodes. As aforementioned, a CRM was utilized in VCN to improve the reconstruction quality. In addition, OctAttention~\cite{Fu2022OctAttentionOL} that presented state-of-the-art  efficiency for the compression of lossy sparse LiDAR PCG is also compared.

Since both OctSqueeze~\cite{Que2021VoxelContextNetAO} and VCN~\cite{Huang2020OctSqueezeOE} do not offer the sources to the public, it is difficult for us to replicate their results. We directly quote the results from VCN paper (i.e., Fig. 6 in~\cite{Que2021VoxelContextNetAO}) and present them in Table~\ref{table:learnedPCC_lidar} and Fig.~\ref{fig:perf_VCN_OctAttention}. As for the OctAttention, we reproduce the results by running its source code. We best ensure the same testing/training conditions as these methods like VCN and OctAttention for comparative study where the details are specified in the supplemental material. 

As seen, our method still achieves leading compression performance when compared with both VCN and OctAttention. For VCN, we also provide its results with the CRM in comparison to  our solution with position offset adjustment. Here we use the same terminology CRM in Table~\ref{table:learnedPCC_lidar} to represent the post-processing module. Clearly, our BD-Rate gains are still retained with CRM. 

Moreover, we show the runtime for all methods when using them to compress LiDAR PCG in Table~\ref{table:learnedPCC_lidar}. Currently, VCN shows the fastest decoding speed. Our method runs about 1$\sim$2 seconds for both encoding and decoding, presenting about 2$\times$ slower than the G-PCC version 14. Because of the causal dependency used in context modeling, OctAttention consumes the longest time for decoding.

\subsection{Discussion} \label{sec:perf_discussion}

Due to various acquisition techniques used in different applications, the characteristics of input point clouds, including the density, volume, precision, noise, etc., vary greatly from one to another. It is extremely challenging to efficiently compress these diverse sources under the same model.
As a compromise, existing solutions apply different models or tools to compress different inputs as aforementioned. 

Previous sections reveal that the SparsePCGC, as a unified model, can compress diverse point clouds in either lossy or lossless mode efficiently. This is mainly because the proposed method can effectively exploit the sparsity nature of unstructured points:

\begin{itemize}
    \item The SparsePCGC deals with the points directly without assuming any prior knowledge of the captured scene, which ideally supports arbitrary-distributed points in the 3D space. The sparse convolution used in this work can significantly reduce the complexity and best leverage valid neighbors within the receptive field to analyze and aggregate local variations;
    \item The multiscale representation leverages the resolution re-scaling for scale-wise construction, with which we can exploit cross-scale correlations to effectively aggregate and embed spatial information from local neighborhood for compression.
\end{itemize}

The proposed SparsePCGC applies model sharing across scales. Lossless SparsePCGC uses the same 8-Stage SOPA model which only consumes 4.9 MB to process either dense or sparse point clouds. The lossy mode of SparsePCGC comprises a lossless and a lossy phase. For the compression of dense PCGs, we use the same 4.9 MB 8-Stage SOPA model in lossless phase,  5.8 MB SLNE enhanced One-Stage SOPA and 0.85 MB One-Stage SOPA in lossy phase; while for the sparse PCGs, the lossless phase requires the 21.7 MB 8-Stage SOPA model since we use a larger kernel size, e.g., 5$\times$5$\times$5, and the lossy phase uses the 31.6 MB SOPA (Position) model. Recently, a number of model compression methods have been developed to reduce the model size~\cite{hong2020efficient}, such as the pruning and quantization techniques, which is an interesting topic for future study.

%% file: table/table_voxeldnn.tex
\begin{table*}[htb]\small
\caption{{Performance comparison of losslessly coded dense point clouds. Compression ratio is measured by the bpp gain {over G-PCC}.}}
\label{table:learnedPCClossless}
\centering
\begin{tabular}{|cccccccc|}
\hline
\multicolumn{1}{|c|}{\multirow{2}{*}{\textbf{Dense PCs}}} & \multicolumn{1}{c|}{\multirow{2}{*}{\textbf{\begin{tabular}[c]{@{}c@{}}G-PCC\\ (seqs/one)\end{tabular}}}} & \multicolumn{1}{c|}{\multirow{2}{*}{\textbf{Ours}}} & \multicolumn{2}{c|}{\textbf{VoxelDNN (one)}~\cite{Nguyen2021LosslessCO, Nguyen2021MultiscaleDC}}         & \multicolumn{2}{c|}{\textbf{NNOC}~\cite{Kaya2021NeuralNM}}                  & \multirow{2}{*}{\textbf{OctAttention}~\cite{Fu2022OctAttentionOL}} \\ \cline{4-7}
\multicolumn{1}{|c|}{}                                    & \multicolumn{1}{c|}{}                                                                                     & \multicolumn{1}{c|}{}                               & \textbf{VoxelDNN} & \multicolumn{1}{c|}{\textbf{MsVoxelDNN}} & \textbf{NNOC} & \multicolumn{1}{c|}{\textbf{fNNOC}} &                                   \\ \hline
\multicolumn{8}{|c|}{\textbf{8iVFB}}                                                                                                                                                                                                                                                                                                                                         \\ \hline
\multicolumn{1}{|c|}{Red\&black$\times$300}                     & \multicolumn{1}{c|}{1.088/1.083}                                                                          & \multicolumn{1}{c|}{0.635}                          & 0.665         & \multicolumn{1}{c|}{0.87}            & 0.723         & \multicolumn{1}{c|}{0.866}          & 0.73                              \\
\multicolumn{1}{|c|}{Loot$\times$300}                           & \multicolumn{1}{c|}{0.961/0.947}                                                                          & \multicolumn{1}{c|}{0.533}                          & 0.577         & \multicolumn{1}{c|}{0.73}            & 0.590          & \multicolumn{1}{c|}{0.743}          & 0.62                              \\
\multicolumn{1}{|c|}{Thaidancer$\times$1}                       & \multicolumn{1}{c|}{0.986}                                                                                & \multicolumn{1}{c|}{0.557}                          & 0.677         & \multicolumn{1}{c|}{0.85}            & 0.684         & \multicolumn{1}{c|}{0.807}          & 0.65                              \\
\multicolumn{1}{|c|}{Boxer$\times$1}                            & \multicolumn{1}{c|}{0.943}                                                                                & \multicolumn{1}{c|}{0.493}                          & 0.550         & \multicolumn{1}{c|}{0.70}            & 0.551         & \multicolumn{1}{c|}{0.682}          & 0.59                              \\ \hline
\multicolumn{1}{|c|}{\textbf{Average}}                    & \multicolumn{1}{c|}{0.994/0.990}                                                                          & \multicolumn{1}{c|}{0.555}                          & 0.617         & \multicolumn{1}{c|}{0.788}           & 0.637         & \multicolumn{1}{c|}{0.775}          & 0.648                             \\
\multicolumn{1}{|c|}{\textbf{Gain}}                       & \multicolumn{1}{c|}{-}                                                                                    & \multicolumn{1}{c|}{\textbf{-44.2\%}}               & -37.7\%       & \multicolumn{1}{c|}{-20.5\%}         & -35.9\%       & \multicolumn{1}{c|}{-22.1\%}        & -34.9\%                           \\ \hline
\multicolumn{1}{|c|}{\textbf{EncTime (s)}}                & \multicolumn{1}{c|}{5.56/5.49}                                                                            & \multicolumn{1}{c|}{2.09}                           & (885)         & \multicolumn{1}{c|}{(54)}              & 91         & \multicolumn{1}{c|}{64}          & \textbf{1.06}                     \\
\multicolumn{1}{|c|}{\textbf{DecTime (s)}}                & \multicolumn{1}{c|}{3.26/3.21}                                                                            & \multicolumn{1}{c|}{\textbf{1.91}}                  & (640)         & \multicolumn{1}{c|}{(58)}              & 1079        & \multicolumn{1}{c|}{66}          & 1229                              \\ \hline
\multicolumn{8}{|c|}{\textbf{MVUB}}                                                                                                                                                                                                                                                                                                                                          \\ \hline
\multicolumn{1}{|c|}{Phil$\times$245}                           & \multicolumn{1}{c|}{1.135/1.142}                                                                          & \multicolumn{1}{c|}{0.714}                          & 0.76          & \multicolumn{1}{c|}{1.02}            & 0.782         & \multicolumn{1}{c|}{1.021}          & 0.79                              \\
\multicolumn{1}{|c|}{Ricardo$\times$216}                        & \multicolumn{1}{c|}{1.061/1.103}                                                                          & \multicolumn{1}{c|}{0.647}                          & 0.687         & \multicolumn{1}{c|}{0.95}            & 0.701         & \multicolumn{1}{c|}{0.941}          & 0.72                              \\ \hline
\multicolumn{1}{|c|}{\textbf{Average}}                    & \multicolumn{1}{c|}{1.098/1.123}                                                                          & \multicolumn{1}{c|}{0.681}                          & 0.724         & \multicolumn{1}{c|}{0.985}           & 0.742         & \multicolumn{1}{c|}{0.981}          & 0.755                             \\
\multicolumn{1}{|c|}{\textbf{Gain}}                       & \multicolumn{1}{c|}{-}                                                                                    & \multicolumn{1}{c|}{\textbf{-37.7\%}}               & -35.5\%       & \multicolumn{1}{c|}{-12.5\%}         & -32.1\%       & \multicolumn{1}{c|}{-10.2\%}        & -30.9\%                           \\ \hline
\multicolumn{1}{|c|}{\textbf{EncTime (s)}}                & \multicolumn{1}{c|}{7.56/9.78}                                                                            & \multicolumn{1}{c|}{2.81}                           & (885)         & \multicolumn{1}{c|}{(85)}              & 101         & \multicolumn{1}{c|}{66}          & \textbf{1.39}                     \\
\multicolumn{1}{|c|}{\textbf{DecTime (s)}}                & \multicolumn{1}{c|}{4.53/5.47}                                                                            & \multicolumn{1}{c|}{\textbf{2.64}}                  & (640)         & \multicolumn{1}{c|}{(92)}              & 1318        & \multicolumn{1}{c|}{69}          & 1520                              \\ \hline
\end{tabular}\\
\vspace{0.05cm}

\end{table*}

%% file: table/table_lossy.tex
\begin{table*}[]\small
\caption{{BD-Rate gains measured using both D1 and D2 for the SparsePCGC against the G-PCC (T) using trisoup codec option, {PCGCv2}~\cite{Wang2020MultiscalePC}, {GeoCNNv2}~\cite{Quach2020ImprovedDP}, and {ADL-PCC}~\cite{Guarda2021AdaptiveDL} for lossy coded dense point clouds.}}
\label{table:BDBR_lossy_dense}
\centering
\begin{tabular}{|c|cc|cc|cc|cc|}
\hline
\multirow{2}{*}{\textbf{Dense PCs}} & \multicolumn{2}{c|}{\textbf{G-PCC (Trisoup)}} & \multicolumn{2}{c|}{\textbf{PCGCv2~\cite{Wang2020MultiscalePC}}} & \multicolumn{2}{c|}{\textbf{ADL-PCC}~\cite{Guarda2021AdaptiveDL}} & \multicolumn{2}{c|}{\textbf{GeoCNNv2}~\cite{Quach2020ImprovedDP}} \\ \cline{2-9} 
                                    & \textbf{D1}           & \textbf{D2}          & \textbf{D1}       & \textbf{D2}      & \textbf{D1}       & \textbf{D2}       & \textbf{D1}        & \textbf{D2}       \\ \hline
longdress\_vox10\_1300              & -76.4\%               & -74.3\%              & -47.2\%           & -36.1\%          & -76.6\%           & -74.7\%           & -74.9\%            & -70.1\%           \\
loot\_vox10\_1200                   & -69.6\%               & -68.0\%              & -45.8\%           & -34.8\%          & -76.4\%           & -74.3\%           & -74.9\%            & -69.6\%           \\
red\&black\_vox10\_1550             & -72.7\%               & -70.6\%              & -45.0\%           & -28.2\%          & -69.6\%           & -68.0\%           & -72.7\%            & -66.2\%           \\
soldier\_vox10\_0690                & -77.4\%               & -76.7\%              & -45.8\%           & -34.7\%          & -72.7\%           & -70.6\%           & -73.1\%            & -68.1\%           \\
queen\_0200                         & -81.8\%               & -80.4\%              & -34.7\%           & -29.2\%          & -77.4\%           & -76.7\%           & -72.5\%            & -66.6\%           \\
player\_vox11\_200                  & -80.6\%               & -78.2\%              & -48.5\%           & -41.9\%          & -81.8\%           & -80.4\%           & -79.5\%            & -75.7\%           \\
dancer\_vox11\_0001                 & -76.4\%               & -74.7\%              & -49.7\%           & -42.8\%          & -80.6\%           & -78.2\%           & -78.4\%            & -73.6\%           \\ \hline
\textbf{Average}                    & \textbf{-89.2\%}      & \textbf{-81.2\%}     & \textbf{-45.2\%}  & \textbf{-35.4\%} & \textbf{-76.4\%}  & \textbf{-74.7\%}  & \textbf{-75.1\%}   & \textbf{-70.0\%}  \\ \hline
\end{tabular}
\end{table*}

%% file: table/table_lidar.tex
\begin{table*}[htb]\small
\caption{BD-Rate gains over the G-PCC anchor for lossily compressed sparse LiDAR PCG. Runtime at the highest bitrate point (12-bit) are exemplified for comparison, all methods are tested on  RTX 2080 GPU.}
\label{table:learnedPCC_lidar}
\centering
\begin{tabular}{|c|cccc|ccc|}
\hline
\multirow{2}{*}{\textbf{KITTI}} & \multicolumn{4}{c|}{\textbf{Test Conditions of VCN}~\cite{Que2021VoxelContextNetAO}}                  & \multicolumn{3}{c|}{\textbf{Test Conditions of OctAttention}~\cite{Fu2022OctAttentionOL}} \\ \cline{2-8} 
                                & \textbf{G-PCC v8} & \textbf{Ours}    & \textbf{VCN} & \textbf{OctSqueeze} & \textbf{G-PCC v14}  & \textbf{Ours}    & \textbf{OctAttention} \\ \hline
\textbf{BD-Rate}                   & -                & \textbf{-25.6\%} & -16.7\%      & -2.1\%           & -                  & \textbf{-22.0\%} & -19.5\%          \\
\textbf{BD-Rate w/ CRM}            & -                & \textbf{-37.7\%} & -30.0\%      & -                & -                  & -                & -                \\ \hline
\textbf{EncTime (s)}            & 1.30             & 1.44             & -      & \textbf{-} & 0.92               & 1.44             & \textbf{0.44}    \\
\textbf{DecTime (s)}            & 0.55             & 1.32             & 0.09         & \textbf{0.008}   & \textbf{0.62}      & 1.19             & 530              \\ \hline
\end{tabular}
\end{table*}

%% file: 5_conclusion.tex
\section{Conclusion And Future Work}
\label{sec:conclusion}

A unified point cloud geometry compression approach is developed, demonstrating state-of-the-art performance in both lossless and lossy compression applications across a variety of datasets, including the dense point clouds (8iVFB, Owlii, MVUB) and the sparse LiDAR point clouds (KITTI, Ford) when compared with the MPEG G-PCC and other popular learning-based approaches.  Moreover, the proposed method attains low {computational consumption} because of the utilization of sparse convolution, and requires a small amount of storage for model parameters, since the same model is shared across scales.

The advantages of the proposed method come from the Multiscale Sparse Tensor Representation, where we use sparse convolutions to directly deal with unstructured points and efficiently aggregate local neighborhood information. We also leverage the multiscale representation to extensively exploit the cross-scale correlations for better context modeling in compression. 

There are numerous interesting topics for exploration in the future, including 1) attribute compression (e.g., RGB colors), 2) motion capturing for dynamic point clouds, and 3) better quality metrics close to human perception  for loss optimization.

A supplemental material is provided with more details about the experimental setup and additional comparisons at \url{ https://github.com/NJUVISION/SparsePCGC/blob/main/Supplementary_Material.pdf}.

\section{Acknowledgement}
Our sincere gratitude is directed to the authors of relevant works used in comparative studies for discussing experiments in detail, and providing the latest results for evaluation.

%% file: 0_pcc.bbl
\begin{thebibliography}{10}
\providecommand{\url}[1]{#1}
\csname url@samestyle\endcsname
\providecommand{\newblock}{\relax}
\providecommand{\bibinfo}[2]{#2}
\providecommand{\BIBentrySTDinterwordspacing}{\spaceskip=0pt\relax}
\providecommand{\BIBentryALTinterwordstretchfactor}{4}
\providecommand{\BIBentryALTinterwordspacing}{\spaceskip=\fontdimen2\font plus
\BIBentryALTinterwordstretchfactor\fontdimen3\font minus
  \fontdimen4\font\relax}
\providecommand{\BIBforeignlanguage}[2]{{%
\expandafter\ifx\csname l@#1\endcsname\relax
\typeout{** WARNING: IEEEtran.bst: No hyphenation pattern has been}%
\typeout{** loaded for the language `#1'. Using the pattern for}%
\typeout{** the default language instead.}%
\else
\language=\csname l@#1\endcsname
\fi
#2}}
\providecommand{\BIBdecl}{\relax}
\BIBdecl

\bibitem{schwarz2019emerging}
S.~Schwarz, M.~Preda, V.~Baroncini \emph{et~al.}, ``Emerging {MPEG} standards
  for point cloud compression,'' \emph{IEEE Journal on Emerging and Selected
  Topics in Circuits and Systems}, vol.~9, pp. 133--148, 2019.

\bibitem{digitalTwin}
H.~X. Nguyen, R.~Trestian, D.~To, and M.~Tatipamula, ``Digital twin for {5G}
  and beyond,'' \emph{IEEE Communications Magazine}, vol.~59, no.~2, pp.
  10--15, 2021.

\bibitem{CABAC}
D.~Marpe, H.~Schwarz, and T.~Wiegand, ``Context-based adaptive binary
  arithmetic coding in the {H.264/AVC} video compression standard,'' \emph{IEEE
  Transactions on Circuits and Systems for Video Technology}, vol.~13, no.~7,
  pp. 620--636, 2003.

\bibitem{InfoTheory_bible}
C.~E. Shannon, ``A mathematical theory of communication,'' \emph{The Bell
  System Technical Journal}, vol.~27, no.~3, pp. 379--423, 1948.

\bibitem{meagher1982geometric}
D.~Meagher, ``Geometric modeling using octree encoding,'' \emph{Computer
  graphics and image processing}, vol.~19, no.~2, pp. 129--147, 1982.

\bibitem{MPEG_PCC_PIEEE}
C.~Cao, M.~Preda, V.~Zakharchenko, E.~S. Jang, and T.~Zaharia, ``Compression of
  sparse and dense dynamic point clouds—methods and standards,''
  \emph{Proceedings of the IEEE}, vol. 109, no.~9, pp. 1537--1558, 2021.

\bibitem{quach2019learning}
M.~Quach, G.~Valenzise, and F.~Dufaux, ``Learning convolutional transforms for
  lossy point cloud geometry compression,'' in \emph{IEEE ICIP}, 2019, pp.
  4320--4324.

\bibitem{Wang2021Lossy}
J.~Wang, H.~Zhu, H.~Liu, and Z.~Ma, ``Lossy point cloud geometry compression
  via end-to-end learning,'' \emph{IEEE Transactions on Circuits and Systems
  for Video Technology}, pp. 1--1, 2021.

\bibitem{Guarda2021AdaptiveDL}
A.~F.~R. Guarda, N.~M.~M. Rodrigues, and F.~Pereira, ``Adaptive deep
  learning-based point cloud geometry coding,'' \emph{IEEE Journal of Selected
  Topics in Signal Processing}, vol.~15, pp. 415--430, 2021.

\bibitem{Nguyen2021MultiscaleDC}
D.~Nguyen, M.~Quach, G.~Valenzise, and P.~Duhamel, ``Multiscale deep context
  modeling for lossless point cloud geometry compression,'' \emph{2021 IEEE
  International Conference on Multimedia Expo Workshops (ICMEW)}, 2021.

\bibitem{Huang2020OctSqueezeOE}
L.~Huang, S.~Wang, K.~Wong, J.~Liu, and R.~Urtasun, ``{OctSqueeze}:
  Octree-structured entropy model for lidar compression,'' \emph{2020 IEEE/CVF
  Conference on Computer Vision and Pattern Recognition (CVPR)}, pp.
  1310--1320, 2020.

\bibitem{NEURIPS2020}
S.~Biswas, J.~Liu, K.~Wong, S.~Wang, and R.~Urtasun, ``{MuSCLE}: Multi sweep
  compression of lidar using deep entropy models,'' in \emph{Advances in Neural
  Information Processing Systems}, vol.~33, 2020.

\bibitem{Que2021VoxelContextNetAO}
Z.~Que, G.~Lu, and D.~Xu, ``{VoxelContext-Net}: An octree based framework for
  point cloud compression,'' \emph{2021 IEEE/CVF Conference on Computer Vision
  and Pattern Recognition (CVPR)}, 2021.

\bibitem{Huang20193DPC}
T.~Huang and Y.~Liu, ``3d point cloud geometry compression on deep learning,''
  \emph{Proceedings of the 27th ACM International Conference on Multimedia},
  2019.

\bibitem{Gao2021PCGCGraphSampling}
L.~Gao, T.~Fan, J.~Wang, Y.~Xu, and Z.~Ma, ``Point cloud geometry compression
  via neural graph sampling,'' \emph{IEEE ICIP}, 2021.

\bibitem{Qi2017PointNetDH}
C.~Qi, L.~Yi, H.~Su, and L.~Guibas, ``{PointNet++}: Deep hierarchical feature
  learning on point sets in a metric space,'' in \emph{NIPS}, 2017.

\bibitem{JPEG2019CTC}
S.~Schwarz, P.~A. Chou, and I.~Sinharoy, ``{JPEG} pleno point cloud coding
  common test conditions,'' \emph{ISO/IEC JTC1/SC29/WG11 N18474}, 2019.

\bibitem{MPEG_LearntPCC}
J.~Pang, M.~A. Lodhi, G.~Martin-Cocher, and D.~Tian, ``{AI}-based point cloud
  compression for new {PCC},'' \emph{ISO/IEC JTC1/SC29/WG11 (MPEG/JPEG)
  m56776}, April 2021.

\bibitem{BMVC2015_150}
B.~Graham, ``Sparse {3D} convolutional neural networks,'' in \emph{BMVC}, 2015.

\bibitem{tmc13code}
``{MPEG-PCC-TMC13},'' \url{https://github.com/MPEGGroup/mpeg-pcc-tmc13},
  accessed: 2021.

\bibitem{BDrate}
G.~Bj{\o}ntegaard, ``Calculation of average {PSNR} differences between
  rd-curves,'' in \emph{ITU-T SG 16/Q6, 13th VCEG Meeting}.\hskip 1em plus
  0.5em minus 0.4em\relax document VCEG-M33, April 2001.

\bibitem{Wang2020MultiscalePC}
J.~Wang, D.~Ding, Z.~Li, and Z.~Ma, ``Multiscale point cloud geometry
  compression,'' \emph{2021 Data Compression Conference (DCC)}, pp. 73--82,
  2021.

\bibitem{8i20178i}
E.~d'Eon, B.~Harrison, T.~Myers, and P.~A.~Chou, ``8i voxelized full bodies - a
  voxelized point cloud dataset,'' \emph{ISO/IEC JTC1/SC29 Joint WG11/WG1
  (MPEG/JPEG) m38673/M72012}, May 2016.

\bibitem{xu2017owlii}
X.~Yi, L.~Yao, and W.~Ziyu, ``Owlii dynamic human mesh sequence dataset,''
  \emph{ISO/IEC JTC1/SC29/WG11 (MPEG/JPEG) m41658}, 2017.

\bibitem{microsoft2019microsoft}
L.~Charles, C.~Qin, O.~Sergio, and A.~C. Philip, ``Microsoft voxelized upper
  bodies - a voxelized point cloud dataset,'' \emph{ISO/IEC JTC1/SC29 Joint
  WG11/WG1 (MPEG/JPEG) m38673/M72012}, May 2016.

\bibitem{Quach2020ImprovedDP}
M.~Quach, G.~Valenzise, and F.~Dufaux, ``Improved deep point cloud geometry
  compression,'' \emph{2020 IEEE MMSP Workshop}, 2020.

\bibitem{Kaya2021NeuralNM}
E.~C. Kaya and I.~Tabus, ``Neural network modeling of probabilities for coding
  the octree representation of point clouds,'' \emph{IEEE MMSP}, 2021.

\bibitem{Fu2022OctAttentionOL}
C.~Fu, G.~Li, R.~Song, W.~Gao, and S.~Liu, ``{OctAttention}: Octree-based
  large-scale contexts model for point cloud compression,'' in \emph{AAAI},
  2022.

\bibitem{Nguyen2021LosslessCO}
D.~Nguyen, M.~Quach, G.~Valenzise, and P.~Duhamel, ``Lossless coding of point
  cloud geometry using a deep generative model,'' \emph{IEEE Transactions on
  Circuits and Systems for Video Technology}, 2021.

\bibitem{Jackins1980Oct}
C.~L. Jackins and S.~L. Tanimoto, ``Oct-trees and their use in representing
  three-dimensional objects,'' \emph{Computer Graphics and Image Processing},
  vol.~14, no.~3, pp. 249--270, 1980.

\bibitem{Schnabel2006Octree}
R.~Schnabel and R.~Klein, ``Octree-based point-cloud compression,'' in
  \emph{Eurographics}, 2006.

\bibitem{Huang2008A}
Y.~Huang, J.~Peng, C.~C. Kuo, and M.~Gopi, ``A generic scheme for progressive
  point cloud coding,'' \emph{IEEE Transactions on Visualization and Computer
  Graphics}, vol.~14, no.~2, pp. 440--453, 2008.

\bibitem{graziosi2020overview}
D.~Graziosi, O.~Nakagami, S.~Kuma, A.~Zaghetto, T.~Suzuki, and A.~Tabatabai,
  ``An overview of ongoing point cloud compression standardization activities:
  video-based {(V-PCC)} and geometry-based {(G-PCC)},'' \emph{APSIPA
  Transactions on Signal and Information Processing}, vol.~9, 2020.

\bibitem{VVC_overview}
B.~Bross, Y.-K. Wang, Y.~Ye, S.~Liu, J.~Chen, G.~J. Sullivan, and J.-R. Ohm,
  ``Overview of the versatile video coding {(VVC)} standard and its
  applications,'' \emph{IEEE Transactions on Circuits and Systems for Video
  Technology}, vol.~31, no.~10, pp. 3736--3764, 2021.

\bibitem{zhu2020view}
W.~Zhu, Z.~Ma, Y.~Xu, L.~Li, and Z.~Li, ``View-dependent dynamic point cloud
  compression,'' \emph{IEEE Transactions on Circuits and Systems for Video
  Technology}, vol.~31, no.~2, pp. 765--781, 2020.

\bibitem{choy20194d}
C.~Choy, J.~Gwak, and S.~Savarese, ``{4D} spatio-temporal convnets: Minkowski
  convolutional neural networks,'' in \emph{IEEE CVPR}, 2019.

\bibitem{ren2018sbnet}
M.~Ren, A.~Pokrovsky, B.~Yang, and R.~Urtasun, ``{SBNet}: Sparse blocks network
  for fast inference,'' in \emph{IEEE CVPR}, 2018, pp. 8711--8720.

\bibitem{Yan2018SECONDSE}
Y.~Yan, Y.~Mao, and B.~Li, ``{SECOND}: Sparsely embedded convolutional
  detection,'' \emph{Sensors (Basel, Switzerland)}, vol.~18, 2018.

\bibitem{SubmanifoldSparseConvNet}
B.~Graham and L.~van~der Maaten, ``Submanifold sparse convolutional networks,''
  \emph{arXiv preprint arXiv:1706.01307}, 2017.

\bibitem{tang2022torchsparse}
H.~Tang, Z.~Liu, X.~Li, Y.~Lin, and S.~Han, ``{TorchSparse}: Efficient point
  cloud inference engine,'' in \emph{Conference on Machine Learning and Systems
  (MLSys)}, 2022.

\bibitem{MinkowskiEngine}
``Minkowskiengine,'' \url{https://github.com/NVIDIA/MinkowskiEngine}, accessed:
  2021.

\bibitem{Szegedy2017Inceptionv4IA}
C.~Szegedy, S.~Ioffe, V.~Vanhoucke, and A.~A. Alemi, ``Inception-v4,
  inception-resnet and the impact of residual connections on learning,'' in
  \emph{AAAI}, 2017.

\bibitem{minnen2018joint}
D.~Minnen, J.~Ball{\'e}, and G.~D. Toderici, ``Joint autoregressive and
  hierarchical priors for learned image compression,'' in \emph{IEEE NeurIPS},
  2018, pp. 10\,771--10\,780.

\bibitem{liu2019non}
T.~Chen, H.~Liu, Z.~Ma, Q.~Shen, X.~Cao, and Y.~Wang, ``End-to-end learnt image
  compression via nonlocal attention optimization and improved contenxt
  modeling,'' \emph{IEEE Trans. Image Processing}, vol.~30, pp. 3179--3191,
  2021.

\bibitem{balle2018variational}
J.~Ball{\'e}, D.~Minnen, S.~Singh \emph{et~al.}, ``Variational image
  compression with a scale hyperprior,'' in \emph{ICLR}, 2018.

\bibitem{Chang2015ShapeNetAI}
A.~X. Chang, T.~Funkhouser, L.~Guibas, P.~Hanrahan \emph{et~al.}, ``Shapenet:
  An information-rich 3d model repository,'' \emph{ArXiv}, vol. abs/1512.03012,
  2015.

\bibitem{Behley2019SemanticKITTIAD}
J.~Behley, M.~Garbade, A.~Milioto \emph{et~al.}, ``{SemanticKITTI}: A dataset
  for semantic scene understanding of lidar sequences,'' \emph{2019 IEEE/CVF
  International Conference on Computer Vision (ICCV)}, pp. 9296--9306, 2019.

\bibitem{mpegdataset}
``{MPEG PCC} dataset,'' \url{http://mpegfs.int-evry.fr/mpegcontent/}, accessed:
  2021.

\bibitem{MPEG_GPCC_CTC}
{WG7, MPEG 3D Graphics Coding}, ``Common test conditions for {G-PCC},''
  \emph{ISO/IEC JTC1/SC29/WG11 N00106}, 2021.

\bibitem{MPEG-AI_PCC}
------, ``Performance analysis of currently {AI}-based available solutions for
  {PCC},'' \emph{ISO/IEC JTC 1/SC 29/WG 7 N0174}, July 2021.

\bibitem{hong2020efficient}
W.~Hong, T.~Chen, M.~Lu, S.~Pu, and Z.~Ma, ``Efficient neural image decoding
  via fixed-point inference,'' \emph{IEEE Transactions on Circuits and Systems
  for Video Technology}, vol.~31, no.~9, pp. 3618--3630, 2020.

\end{thebibliography}
